\documentclass[nohyperref]{article}

\usepackage[accepted]{icml2022}

\usepackage{adjustbox}
\usepackage{amsmath}
\usepackage{amssymb}
\usepackage{amsthm}
\usepackage{array}
\usepackage{bm}
\usepackage{booktabs}
\usepackage{cite}
\usepackage[bottom,flushmargin]{footmisc}
\usepackage{graphicx}
\usepackage{hyperref}
\usepackage{makecell}
\usepackage{mathtools}
\usepackage{microtype}
\usepackage{multirow}
\usepackage{pifont}
\usepackage{setspace}
\usepackage[small]{caption}
\usepackage{scalerel}
\usepackage{soul}
\usepackage{subcaption}
\usepackage{widetable}

\newcommand{\squeeze}[1]{{#1\parfillskip=0pt\par}}
\newcommand{\pix}{\kern 0.1em}

\usepackage[capitalize,noabbrev]{cleveref}
\usepackage{listings}
\definecolor{keywordstyle}{RGB}{207,35,47}
\definecolor{backgroundstyle}{RGB}{247,249,250}
\definecolor{stringstyle}{RGB}{11,48,104}
\definecolor{commentstyle}{RGB}{110,118,128}
\definecolor{identifierstyle}{RGB}{37,40,47}
\definecolor{basicstyle}{RGB}{27,95,181}
\definecolor{customkeystyle}{RGB}{148,57,0}
\definecolor{functionstyle}{RGB}{130,80,223}

\usepackage{csvsimple}
\usepackage[T1]{fontenc}

\usepackage[normalem]{ulem}
\usepackage{contour}
\usepackage{ulem}

\contourlength{0.6pt}

\let\OLDthebibliography\thebibliography
\renewcommand\thebibliography[1]{
\OLDthebibliography{#1}
\setlength{\itemsep}{5pt plus 2pt minus 1.5pt}
}

\usepackage[algoruled,vlined,nofillcomment,algo2e]{algorithm2e}
\SetKwComment{Comment}{$\triangleright$}{}
\DontPrintSemicolon
\usepackage{enumitem}
\newenvironment{protocol}[1][htb]
{
 \begin{algorithm2e}[#1]%
}{\end{algorithm2e}}

\usepackage{etoolbox}
\providetoggle{beginFlag}
\settoggle{beginFlag}{true}

\SetAlgoSkip{SkipBeforeAndAfter}

\icmltitlerunning{HyperImpute}
\begin{document}

\twocolumn[
\icmltitle{HyperImpute:\\Generalized Iterative Imputation with Automatic Model Selection}

\icmlsetsymbol{equal}{*}

\begin{icmlauthorlist}
\icmlauthor{Daniel Jarrett}{equal,cam}
\icmlauthor{Bogdan Cebere}{equal,cam}
\icmlauthor{Tennison Liu}{cam}
\icmlauthor{Alicia Curth}{cam}
\icmlauthor{Mihaela van der Schaar}{cam,ucla}
\end{icmlauthorlist}

\icmlaffiliation{cam}{Department of Applied Mathematics \& Theor- etical Physics, University of Cambridge, UK.}
\icmlaffiliation{ucla}{Department of Elec- trical Engineering, University of California, Los Angeles, USA}

\icmlcorrespondingauthor{Bogdan Cebere}{bcc38@cam.ac.uk}

\icmlkeywords{Machine Learning, ICML}

\vskip 0.185in
]

\printAffiliationsAndNotice{\icmlEqualContribution}

\begin{abstract}

\squeeze{
Consider the problem of imputing missing values in a dataset.
On the one hand, \mbox{conventional app-} roaches using iterative imputation benefit from the simplicity and customizability of learning conditional distributions directly, but suffer from the practical requirement for appropriate model specification of each and every variable.
On the other hand, recent methods using deep generative modeling benefit from the capacity and efficiency of learning with neural network function approximators, but are often difficult to optimize and rely on stronger data assumptions.
In this work, we~study an approach that marries the advantages of both:
We propose \textit{HyperImpute}, a generalized iterative imputation framework for adaptively and automatically configuring column-wise models and their hyperparameters.
Practically, we provide a concrete implementation with out-of-the-box learners, optimizers, simulators, and extensible interfaces.
Empirically, we investigate this framework via comprehensive experiments and sensitivities on a variety of public datasets, and demonstrate its ability to generate accurate imputations relative to a strong suite of benchmarks.
Contrary to re-
cent work, we believe our findings constitute a str- ong defense of the iterative imputation paradigm.}
\end{abstract}
\vspace{-1.65em}

\begin{center}
\href{https://github.com/vanderschaarlab/hyperimpute}{https://github.com/vanderschaarlab/hyperimpute}.
\end{center}
\vspace{-0.85em}

\section{Introduction}\label{introduction}

\squeeze{
Missing data is a ubiquitous problem in real-life data collection. For instance, certain characteristics of a patient may not have been recorded properly during their visit, but we may nevertheless be interested in knowing the values those variables most likely took on \cite{barnard1999applications,mackinnon2010use,stekhoven2012missforest,alaa2017personalized}. Here, we consider precisely this problem of \textit{imputing} the missing values in a dataset. Specifically, consider the general case where different records in a dataset may contain missing values for different variables, so no columns are assumed to be complete.
}

\squeeze{
Most popular approaches fall into two main categories. On the one hand, conventional approaches using \textit{iterative imputation} operate by estimating the conditional distributions of each feature on the basis of all the others, and missing values are imputed using such univariate models in a round-robin fashion until convergence \cite{brand1999development,heckerman2000dependency,raghunathan2001multivariate,gelman2004parameterization,van2006fully,van2011mice}. In theory, this approach affords great customizability in creating multivariate models: by simply specifying univariate models, one can easily and implicitly work with joint models outside any known parametric multivariate density \cite{van2011mice,liu2014stationary,zhu2015convergence}.
In practice, however, this strategy often suffers from the requirement that models for every single variable be properly specified. In particular, for each column with missing values, one needs to choose the functional form for the model, select the set of regressors as input, include any interaction terms of interest, add appropriate regularization, or handle derived variables separately to prevent collinearity in the common case of linear models; this is time-consuming and dependent on human expertise.
}

\squeeze{
On the other hand, recent methods using \textit{deep generative models} operate by estimating a joint model of all features together, from which missing values can be queried \cite{rezende2014stochastic,gondara2017multiple,mattei2018leveraging,yoon2018gain,li2019misgan,yoon2020gamin,nazabal2020handling,ivanov2018variational,richardson2020mcflow,mattei2019miwae}. In theory, these approaches more readily take advantage of the capacity and efficiency of learning using deep function approximators and the ability to capture correlations among covariates by amortizing the parameters \cite{nazabal2020handling,goodfellow2016deep}.
In practice, however, this strategy comes at the price of much more challenging optimization---GAN-based imputers \cite{yoon2018gain,li2019misgan,yoon2020gamin,dai2021multiple} are often prone to the usual difficulties in adversarial training \cite{salimans2016improved,thanh2020catastrophic}, and VAE-based imputers \cite{nazabal2020handling,ivanov2018variational} are subject to the usual limitations of training latent-variable models through variational bounds \cite{zhao2017towards,lucas2019understanding}; empirically, these methods may often be outperformed by iterative imputation \cite{muzellec2020missing,jager2021benchmark}. Further, most of such techniques---with the notable exception to \cite{mattei2019miwae}---either separately require fully-observed datasets during training \cite{rezende2014stochastic,gondara2017multiple,mattei2018leveraging}, or operate on the strong assumption that missingness patterns are entirely independent of both observed and unobserved data \cite{yoon2018gain,li2019misgan,yoon2020gamin,nazabal2020handling,ivanov2018variational,richardson2020mcflow}, which is not realistic.
}

\begin{table*}[t]\small
\newcolumntype{A}{>{          \arraybackslash}m{4.5 cm}}
\newcolumntype{B}{>{\centering\arraybackslash}m{2.1 cm}}
\newcolumntype{C}{>{\centering\arraybackslash}m{2.5 cm}}
\newcolumntype{D}{>{\centering\arraybackslash}m{2.5 cm}}
\newcolumntype{E}{>{\centering\arraybackslash}m{2.5 cm}}
\newcolumntype{F}{>{\centering\arraybackslash}m{2.1 cm}}
\newcolumntype{G}{>{\centering\arraybackslash}m{2.1 cm}}
\setlength{\cmidrulewidth}{0.5pt}
\setlength\tabcolsep{0pt}
\renewcommand{\arraystretch}{0.99}
\vspace{0.15em}
\caption{\squeeze{\textit{Comparison with Related Work}. $^{\text{1}}$Denotes the most general regime under which each method is appropriate. (But note that methods are often empirically tested in all three regimes, not necessarily with theoretical motivation). $^{\text{2}}$Except MIWAE \cite{mattei2019miwae}. $^{\text{3}}$Only for HI-VAE \cite{nazabal2020handling}.}}
\vspace{-1.25em}
\label{tab:related}
\begin{center}
\begin{adjustbox}{max width=\textwidth}
\begin{tabular}{ABCDEFG}
\toprule
  \textbf{Technique}
& \textbf{Examples}
& \textbf{Missing Pattern}$^{\text{1}}$
& \textbf{Required Data}
& \textbf{Data Types}
& \textbf{Column-wise}
& \textbf{Auto Selection}
\\
\midrule
  Mean Imputation
& \cite{hawthorne2005imputing}
& MCAR-only
& Incomplete
& Continuous
& -
& -
\\
  Discriminative, 1-Shot
& \cite{silva2011missing,euredit2005interim}
& MCAR-only
& Fully-Observed
& Mixed
& Yes
& No
\\
  Discriminative,~\pix Iterative
& \cite{stekhoven2012missforest,brand1999development,heckerman2000dependency,raghunathan2001multivariate,gelman2004parameterization,van2006fully}
& MAR
& Incomplete
& Mixed
& Yes
& No
\\
  Generative, Implicit
& \cite{yoon2018gain,li2019misgan,yoon2020gamin}
& MCAR-only
& Incomplete
& Mixed
& No
& No
\\
  Gen., Explicit (Full Input)
& \cite{rezende2014stochastic,gondara2017multiple,mattei2018leveraging}
& MAR
& Fully-Observed
& Continuous
& No
& No
\\
  Gen., Explicit (Incomplete Input)
& \cite{nazabal2020handling,ivanov2018variational,richardson2020mcflow,mattei2019miwae}
& ~\pix MCAR-only$^{\text{2}}$
& Incomplete
& Mixed
& ~\pix Yes$^{\text{3}}$
& No
\\
  Optimal Transport
& \cite{muzellec2020missing}
& MAR
& Incomplete
& Mixed
& No
& No
\\
\midrule
  \textbf{HyperImpute}
& \textbf{(Ours)}
& \textbf{MAR}
& \textbf{Incomplete}
& \textbf{Mixed}
& \textbf{Yes}
& \textbf{Yes}
\\
\bottomrule
\end{tabular}
\end{adjustbox}
\end{center}
\vspace{-1.75em}
\end{table*}

\setcounter{footnote}{3}

\squeeze{
\textbf{Three Desiderata}~~
Can we do better? In light of the preced- ing discussion, we argue that a good baseline solution to the imputation problem should satisfy the following criteria:}
\vspace{-1.25em}

\begin{itemize}[leftmargin=1em]
\itemsep-2pt
\item \squeeze{\textbf{\textit{Flexibility}}: It should combine the flexibility of conditio- nal specification with the capacity of deep approximators.}
\item \squeeze{\textbf{\textit{Optimization}}: It should relieve the burden of complete specification, and be easily and automatically optimized.}
\item \squeeze{\textbf{\textit{Assumptions}}: It should be trainable without complete data, but not assume missingness is completely random.}
\end{itemize}
\vspace{-0.75em}

\squeeze{
\textbf{Contributions}~
In this work, we present a simple but effecti- ve method that satisfies these criteria, facilitates accessibility and reproducibility in imputation research, and constitutes a strong defense of the iterative imputation paradigm.
Our contributions are three-fold.
First, we formalize the imputation problem and describe \textit{HyperImpute}, a generalized iterative framework for adaptively and automatically configuring column-wise models and their hyperparameters \mbox{(Section \ref{three})}.
Second, we give a practical implementation with out-of-the-box learners, optimizers, simulators, and extensible interfaces (Section \ref{method}).
Third, we empirically investigate this me- thod via comprehensive experiments and sensitivities, and demonstrate its ability to generate accurate imputations relative to strong benchmarks (Section \ref{experiments}). Contrary to what recent work suggests, we find that iterative imputation---done right---consistently outperforms more complex alternatives.
}

\vspace{-0.325em}
\section{Background}~\label{background}
\vspace{-1.5em}

\squeeze{
By way of preface, two key distinctions warrant emphasis:
First, we are focusing on imputing missing values \textit{as~an~end} in and of itself---that is, to estimate
what those values probably looked like.
In particular, we are not focusing on imputing missing values \textit{as a means} to obtain input for some known downstream task---such as regression models for predicting labels \cite{morvan2020neumiss,perez2022benchmarking}, generative models for synthetic data \cite{li2019misgan,ma2020vaem}, or active sensing models for information acquisition \cite{ma2018eddi,ma2020vaem,lewis2021accurate}; these motivate concerns fundamentally entangled with each downstream task, and often call for joint training to directly minimize the objectives of those end goals \cite{morvan2021s,li2019misgan}.~Here, we focus solely on the imputation problem itself.
}

\squeeze{
Second, we restrict our discussion to the (most commonly studied) setting where missingness patterns depend only on the \textit{observed} components of the data, and not the \textit{missing} components themselves. Briefly, data may be classified as ``missing completely at random'' (MCAR), where the missingness does not depend on the data at all; ``missing at random'' (MAR), where the missingness depends only on the observed components; or ``missing not at random'' (MNAR), where the missingness depends on the missing components themselves as well \cite{rubin1976inference,rubin1987multiple,van2018flexible,seaman2013meant}.
(These notions are formalized mathematically in Section \ref{three}).
In MCAR and MAR settings, the non-response is ``ignorable'' in the sense that inferences do not require modeling the missingness mechanism itself \cite{van2018flexible,seaman2013meant}. This is not the case in the MNAR setting, where the missing data distribution is generally impossible to identify without imposing domain-specific assumptions, constraints, or parametric forms for the missingness mechanism \cite{jung2011latent,little1993pattern,mohan2018estimation,sportisse2020estimation,kyono2021miracle,ipsen2020not,ma2021identifiable}. Here, we limit our attention to MCAR and MAR settings.
}

\squeeze{
\textbf{Related Work}~
Table \ref{tab:related} presents relevant work in our setting, and summarizes the key advantages of HyperImpute over prevailing techniques. State-of-the-art methods can be categorized as discriminative or generative. In the former, \textit{iterative} methods are the most popular, and consist in specifying a univariate model for each feature on the basis of all others, and performing a sequence of regressions by cycling through each such target variable until all models converge; well-known examples include the seminal MissForest and MICE, and their imputations are valid in the MAR setting \cite{van2011mice,stekhoven2012missforest,van2018flexible}. Less effectively, \textit{one-shot} methods first train regression models using fully-observed training data, which are then applied to incomplete testing data for imputation, and are only appropriate under the more limited MCAR setting \cite{silva2011missing,euredit2005interim}.
}

\squeeze{
On the generative side, \textit{implicit} models consist of imputers trained as generators in GAN-based frameworks; despite their popularity, imputations they produce are only valid under the MCAR assumption \cite{yoon2018gain,li2019misgan,yoon2020gamin}. Alternatively, \textit{explicit} models refer to deep latent-variable models trained to approximate joint densities using variational bounds; as noted in Section \ref{introduction}, most either rely on having fully-observed training data \cite{rezende2014stochastic,gondara2017multiple,mattei2018leveraging}, or otherwise are only appropriate for use under the MCAR assumption  \cite{nazabal2020handling,ivanov2018variational,richardson2020mcflow}. The only exception is MIWAE \cite{mattei2019miwae}, which adapts the objective of importance-weighted autoencoders \cite{burda2015importance} to approximate maximum likelihood in MAR settings; but their bound is only tight in the limit of infinite computational power. Further, save methods that fit separate decoders for each feature \cite{nazabal2020handling}, generative methods do not accommodate column-specific modeling.
}

\squeeze{
Finally, for completeness there are also traditional methods based on
mean substitution \cite{hawthorne2005imputing},
hot deck imputation \cite{marker2002large},
\textit{k}-nearest neighbors \cite{troyanskaya2001missing},
EM-based joint models \cite{garcia2010pattern},
matrix completion using low-rank assumptions \cite{hastie2015matrix}, as well as
a recently proposed technique based on optimal transport \cite{muzellec2020missing}.
}

\vspace{-0.325em}
\section{HyperImpute}~\label{three}
\vspace{-1.5em}

\squeeze{
We begin by formalizing our imputation problem and setting (Section \ref{problem_formulation}). Motivated by our three criteria, we propose performing generalized iterative imputation \mbox{(Section \ref{gii})}, which we solve by automatic model selection \mbox{(Section \ref{ams})}, yielding our proposed \textit{HyperImpute} algorithm \mbox{(Section \ref{hia})}.}

\vspace{-0.35em}
\subsection{Problem Formulation}\label{problem_formulation}

\squeeze{
Let \smash{$\mathbf{X}:=(X_1,..., X_D) \in \mathcal{X}=\mathcal{X}_1\times...\times\mathcal{X}_D$} denote a $D$-dimensional random variable, where \smash{$\mathcal{X}_d\subseteq\mathbb{R}$}
(continuous), and \smash{$\mathcal{X}_d = \{1,..., K_d\}$} (categorical), for $d\in$
$\{1,...,D\}$. We shall adopt notation similar to recent work (see e.g. \cite{yoon2018gain,seaman2013meant}).
}

\textbf{Incomplete Data}~
We do not have \textit{complete} observational access to $\mathbf{X}$; instead, access is mediated by random masks \smash{$\mathbf{M}:=(M_{1},...,M_{D})\in\{0,1\}^{D}$}, such that $X_{d}$ is observable precisely when $M_{d}=1$. Formally, let \smash{$\tilde{\mathcal{X}}_{d}:=\mathcal{X}_{d}\cup\{*\}$} aug- ment the space for each $d$, where ``$*$'' denotes an unobserved value. Then the \textit{incomplete} random variable that we observe is given by $\tilde{\mathbf{X}}:=(\tilde{X}_1,..., \tilde{X}_D)\in\tilde{\mathcal{X}}=\tilde{\mathcal{X}}_1\times...\times\tilde{\mathcal{X}}_D$, with

\vspace{-2.25em}
\begin{align}\label{corrupt}
\tilde{X}_d:=
\begin{cases}
X_{d},              & \text{if~} M_{d}=1\\[-0.5ex]
~\pix*\hspace{4pt}, & \text{if~} M_{d}=0
\end{cases}
\end{align}
\vspace{-1.75em}

\textbf{Imputation Problem}~
Suppose we are given an (incomplete) dataset \smash{$\mathcal{D}$\pix$:=$\pix$\{(\tilde{\mathbf{X}}^n,\mathbf{M}^n)\}^N_{n=1}$} of $N$ records. (In the sequel, we shall drop indices $n$ unless otherwise necessary). We wish to \textit{impute} the missing values for any and all records \smash{$\tilde{\mathbf{X}}$}---that is, to approximately reverse the corruption process of Equation \ref{corrupt} by generating $\hat{\mathbf{X}}:=(\hat{X}_1,..., \hat{X}_D)\in\mathcal{X}$. For each $d$ with $M_{d}=0$, let $\bar{X}_{d}$ denote its imputed value. Then

\vspace{-2.25em}
\begin{align}
\hat{X}_d:=
\begin{cases}
\tilde{X}_{d}, & \text{if~} M_{d}=1\\[-0.5ex]
\bar{X}_{d},   & \text{if~} M_{d}=0
\end{cases}
\end{align}
\vspace{-1.75em}

\squeeze{
\textbf{Missingness Mechanism}~
Let \smash{$S_{\mathbf{M}}:=\{d:M_{d}=1\}$} be the set of indices picked out by $\mathbf{M}$, and \mbox{define the} \textit{selector} projection \smash{$f_{\mathbf{M}}:\mathbf{X}\mapsto f_{\mathbf{M}}(\mathbf{X})=(X_{d})_{d\in S_{\mathbf{M}}}$} from $\mathcal{X}$ onto the subspace $\raisebox{-0.5pt}{\scalebox{1.15}{$\Pi$}}_{d\in S_{\mathbf{M}}}\mathcal{X}_{d}$. Note that $f_{\mathbf{M}}$ induces a \textit{partition} of $\mathbf{X}$~$=$~$(\mathbf{X}_{\text{obs}},\mathbf{X}_{\text{mis}})$, where $\mathbf{X}_{\text{obs}}:=f_{\mathbf{M}}(\mathbf{X})$ is the observed component and $\mathbf{X}_{\text{mis}}:=f_{\mathbf{1}-\mathbf{M}}(\mathbf{X})$ the missing component. The \textit{missingness mechanism} for the incomplete variable $\tilde{\mathbf{X}}$ is
}

\vspace{-1.75em}
\begin{equation}
{\arraycolsep=3pt
~~~~~\begin{array}{lll}
  \text{MCAR},
& \text{if}~\forall\mathbf{M},\mathbf{X},\mathbf{X}':
& ~p(\mathbf{M}|\mathbf{X})=p(\mathbf{M}|\mathbf{X}')\\
  \multirow{2}{*}{$\text{MAR}\hspace{7pt},$}
& \multirow{2}{*}{$\text{if}~\forall\mathbf{M},\mathbf{X},\mathbf{X}':$}
& ~\mathbf{X}_{\text{obs}}=\mathbf{X}'\hspace{-3pt}\raisebox{0pt}{$_{\text{obs}}$}
\Rightarrow\\
&
& ~p(\mathbf{M}|\mathbf{X})=p(\mathbf{M}|\mathbf{X}')\\
  \text{MNAR},
& \multicolumn{2}{l}{\text{if neither of the above conditions hold.}}
\end{array}
}
\end{equation}
\vspace{-1.5em}

\squeeze{
Throughout, we assume our data $\mathcal{D}$ is MCAR or MAR.
While recent literature often uses similar classifications in discussion \cite{yoon2018gain,yoon2020gamin,nazabal2020handling,mattei2019miwae,muzellec2020missing}, most are not rigorously scoped}

\squeeze{
or otherwise use definitions that have been shown to be ambiguous (see e.g. discussion of \cite{seaman2013meant} on \cite{van2011mice,van2018flexible,little2002statistical,fitzmaurice2012applied}).         }

\vspace{-0.35em}
\subsection{Generalized Iterative Imputation}\label{gii}

\squeeze{
Recall our criteria from Section \ref{introduction}: Consider parsimony in \textit{assumptions} (viz. criterion 3): Neither do we wish to assume complete data for training, nor assume the data is MCAR. This immediately rules out most of Table~\ref{tab:related}, leaving only iterative imputation (e.g. MICE \cite{van2011mice}), deep generation (i.e. MIWAE \cite{mattei2019miwae}), and optimal transport (i.e. Sinkhorn \cite{muzellec2020missing}).
Next, consider \textit{flexibility} (viz. criterion 1): Only iterative imputation allows specifying different models for each feature, and is important in practice: Conditional specifications span a much larger space than the space of known joint models, and uniquely permit incorporating design-specific considerations such as bounds and interactions---difficult to do so with a single joint density, parametric or otherwise \cite{van2011mice,gelman2004parameterization,zhu2015convergence}.
}

\squeeze{
Let $\mathcal{A}$ be some space of univariate models and hyperparameters. Classic iterative imputation requires a specification
$a_{d}\in\mathcal{A}$ for each column (e.g. linear regression), with corresponding hypothesis space $\mathcal{H}_{d}$ (e.g. regression coefficients). Now, it is known that conditionally-specified models may not always induce valid joint distributions \cite{gelman2004parameterization,berti2014compatibility}, and that poorly-fitting conditional models may lead to biased results \cite{drechsler2008does,zhu2015convergence}. Let $h_{d}\in\mathcal{H}_{d}$ be a hypothesis for the $d$-th model:
}

\vspace{-1.15em}
\begin{equation}
p(X_{d}|X_{1},...,X_{d-1},X_{d+1},...,X_{D};h_{d})
\end{equation}
\vspace{-1.85em}

\squeeze{
and let \smash{$\mathcal{H}_{\text{com}}$} be the space of tuples $(h_{1},...,h_{D})$ that induce valid joint distributions. Augmenting the capacity of univari- ate models makes it more likely \smash{$\mathcal{H}_{\text{com}}\subset\raisebox{-0.5pt}{\scalebox{1.15}{$\Pi$}}_{d=1}^{D}\mathcal{H}_{d}$} so that the true joint distribution is embedded in the parameter space of the conditionals---thereby improving results.\footnote{We defer to treatment in \cite{arnold2001conditionally,gelman2004parameterization,liu2014stationary,zhu2015convergence,van2018flexible} for detailed discussion of consistency and convergence properties of iterative imputation.\vspace{-0.35em}}
So as our first step, we propose to generalize the iterative method beyond learning hypotheses $h_{d}\in\mathcal{H}_{d}$ that correspond to a \textit{specific} $a_{d}$, but instead to search over \textit{all models and hyperparameters} in $\mathcal{A}$ itself---which allows us to incorporate the capacity of state-of-the-art function approximators such as deep neural networks and modern boosting techniques.
}

\vspace{-0.45em}
\subsection{Automatic Model Selection}\label{ams}

\squeeze{
Prima facie, we have only made things harder w.r.t. \textit{optimization} (viz. criterion 2): We have now added the complexity of multiple flexible classes of learners and their hyperparameters.
Choosing the best set of specifications is highly non-trivial: It depends on the characteristics of each feature, the relationships among them, the number of training samples, feature dimensionalities, and the rates and patterns of data missingness.
However, this has received little attention in related work: The burden is often placed on the user, for whom domain expertise is assumed sufficient \cite{brand1999development,heckerman2000dependency,raghunathan2001multivariate,gelman2004parameterization,van2006fully}. Instead, can these be \textit{automatically} selected, configured, and optimized?
}

\squeeze{
In leveraging AutoML \cite{hutter2019automated,he2021automl} to the rescue, we first consider the standard ``top-down'' search strategy (i.e. with a single global optimizer; see e.g. \cite{wang2017batched,alaa2018autoprognosis} as applied in practice). In the following, let $A$ denote the cardinality of the space of models and hyperparameters (for each univariate model), let $K$ denote an upper bound on the number of iterations for iterative procedure to converge (under any specification), and recall that $D$ denotes the number of feature dimensions.}
\vspace{-0.75em}
\begin{itemize}[leftmargin=1em]
\itemsep-0pt
\item \textbf{Top-Down Search}:
Search over the entire space of \textit{combinations} of univariate models and hyperparameters. So, the iterative imputation procedure (run to completion) is called within a global search loop. The size of the search space is $A^{D}$, and each evaluation calls the iterative procedure once, which runs $O(KD)$ regressions. For search algorithms that reduce complexity by a constant factor, the overall complexity of the optimization is $O(KDA^D)$.
\item \squeeze{\textbf{Concurrent Search}:
What if we optimized all columns in parallel, on a \textit{per-column} basis? This can be done by repeatedly calling iterative imputation (run to completion) within a global search loop, but evaluating and optimizing within each univariate search space independently of others. The size of each search space is $A$, and as before each search evaluation requires $O(KD)$\pix regressions. Under the same assumptions, the overall complexity is $O(KDA)$.}
\end{itemize}
\vspace{-0.75em}
\squeeze{
Computationally, the former is clearly intractable. But the latter is also undesirable, since each univariate model is optimized on its own---and is thus unaware of how the models and hyperparameters being selected for the other columns may potentially affect the best model and hyperparameters for the current column. Can we do better? Instead of calling an (inner) iterative procedure within an (outer) search procedure, we propose an ``inverted'' strategy that calls an (inner) search procedure within an (outer) iterative procedure:
}
\vspace{-0.75em}
\begin{itemize}[leftmargin=1em]
\item \squeeze{\textbf{Inside-Out Search}:
Begin with an iterative imputation procedure, which runs $O(KD)$ regressions. Each regression is carried out by searching over the space of univariate models and hyperparameters, for which the size of the search space is $A$. The overall iterative imputation procedure is run to completion. For search algorithms that reduce complexity by a constant factor, the overall complexity is $O(KDA)$. We call this strategy \textit{HyperImpute}.}
\end{itemize}
\vspace{-0.75em}
\squeeze{
On the one hand, performing the iterative loop on the outside allows us to inherit the usual properties of classic iterative imputation---specifically, that (i.) the imputations are valid in general under the MAR assumption \cite{van2006fully,van2011mice}; that (ii.) they asymptotically pull toward the consistent model when the joint distribution is realizable in the parameter space of the conditional specifications \cite{liu2014stationary,zhu2015convergence}; and that (iii.) concerns about incompatibility or non-convergence are seldom serious in practice \cite{drechsler2008does,van2018flexible}. On the other hand, performing the search procedure on the inside allows us to benefit from automatic model selection among flexible function approxi-
mators and their hyperparameters, while obtaining the same lower complexity of the (more naive) concurrent strategy.
}

\vspace{-0.65em}
\subsection{The HyperImpute Algorithm}\label{hia}
\vspace{0.65em}

\begin{protocol}[h!]\small
\caption{HyperImpute}
\label{alg:hyperimpute}
\begin{spacing}{1.0}
\SetKwInput{KwInputs}{Input}
\SetKwInput{KwOutput}{Output}
\SetKwInput{KwParameters}{Parameters}
\SetKwInput{KwInit}{Initialize}
\SetKw{KwIn}{in}
\SetKwFunction{FColSearch}{ModelSearch}
\SetKwFunction{InitImpute}{BaselineImpute}
\KwParameters{\mbox{Global set of models \& hyperparameters $\mathcal{A}$},\\
\mbox{{\FColSearch} function, {\InitImpute} function,}\\
\mbox{Imputation stop criterion $\gamma$, Selection skip criterion $\sigma$,} \\ Column visitation order $\pi$}
\KwInputs{Incomplete dataset \smash{$\mathcal{D}:=\{(\tilde{\mathbf{X}}^n,\mathbf{M}^n)\}^N_{n=1}$}}
\KwOutput{Imputed dataset \smash{$\hat{\mathcal{D}}:=\{\hat{\mathbf{X}}^n\}^N_{n=1}$}}
\KwInit{\smash{\raisebox{-1pt}{$\hat{\mathcal{D}}$}~$\leftarrow$~\InitImpute{$\mathcal{D}$}}}
\While(\Comment*{\small\textnormal{~keep imputing?\hspace{-13pt}}}\vspace{-1.1em}){\upshape $\gamma$ is False}{
\For{\upshape column $d\in$ visitation order $\pi$}{
$\hat{\mathcal{D}}_{-d}^{\pix\text{obs}}:=\{\hat{\mathbf{X}}_{-d}^n\}_{n: M^n_d=1}$\hfill\mbox{(5)\hspace{-12.5pt}}
\\
$\mathcal{D}_{d}^{\text{obs}}:=\{\hat{X}^n_d\}_{n: M^n_d=1}$\hfill\mbox{(6)\hspace{-12.5pt}}
\\
\If(\Comment*{\small\textnormal{~keep selecting?\hspace{-12.5pt}}}\vspace{-1.1em}){\upshape $\sigma$ is False}{
$a_d\leftarrow$~\FColSearch{$\mathcal{D}_d^{\text{\upshape obs}},\hat{\mathcal{D}}_{-d}^{\pix\text{\upshape obs}},\mathcal{A}$}\\ }
\vspace{-0.3em}$h_{d}\leftarrow a_d$\texttt{.train(}$\mathcal{D}_d^{\text{obs}},\hat{\mathcal{D}}_{-d}^{\pix\text{obs}},\mathcal{H}_{d}$\texttt{)}\vspace{0.1em}\\
$\hat{\mathcal{D}}_{-d}^{\pix\text{mis}}:=\{\hat{\mathbf{X}}_{-d}^n\}_{n: M^n_d=0}$\hfill\mbox{(7)\hspace{-12.5pt}}
\\
\mbox{$\hat{\mathcal{D}}_{d}^{\text{mis}}:=\{\hat{X}^n_d\}_{n: M^n_d=0}\leftarrow h_d$\texttt{.impute(}$\hat{\mathcal{D}}_{-d}^{\pix\text{mis}}$\texttt{)}}\mbox{(8)\hspace{-12.5pt}}
\\
}
\Return{$\hat{\mathcal{D}}$}
}
\end{spacing}
\end{protocol}

\vspace{-0.35em}

\squeeze{
Algorithm \ref{alg:hyperimpute} presents HyperImpute. We begin by performing a \textit{baseline imputation}, such as simple mean substitution. Next, we iterate through the dataset column by column, refining the imputations for each feature as we go. Specifically, at each iteration (i.e. for each feature dimension $d$), we first locate all records for which that feature is observed, and collect the \textit{observed target} $\mathcal{D}_{d}^{\text{obs}}$ and its regressors $\hat{\mathcal{D}}_{-d}^{\pix\text{obs}}$. (Note that some components of the regressors may themselves be imputations). We perform \textit{model selection} to find the best model (and hyperparameter) $a_{d}\in\mathcal{A}$ for the observed target. This is used to learn a hypothesis $h_{d}\in\mathcal{H}_{d}$ for imputing the target value of all records for which that feature is missing, i.e. the \textit{missing target} $\hat{\mathcal{D}}_{d}^{\text{mis}}$, on the basis of regressors $\hat{\mathcal{D}}_{-d}^{\pix\text{mis}}$.\footnote{\squeeze{\footnotesize The \textit{target} \smash{$\mathcal{D}_{d}^{\text{obs}}$$:=$}\smash{\footnotesize$\{\hat{X}^n_d\}_{n: M^n_d=1}$} (6) contains the entries of that column ($d$) whose value is not missing (\smash{$M^n_d=1$}).
The \textit{regressors} \smash{$\hat{\mathcal{D}}_{-d}^{\pix\text{obs}}$$:=$$\{\hat{\mathbf{X}}_{-d}^n\}_{n: M^n_d=1}$}~(5) contains all other columns ($-d$) for those same rows. Importantly, we should be clear that the ``\smash{$M^n_d$\pix$=$\pix$1$}'' is picking out rows where \textit{targets} (not \textit{regressors}!) are observed.}\vspace{-1.15em}}}

\squeeze{
The outer procedure is performed until an \textit{imputation stopping criterion} $\gamma$ is met, e.g. based on the incremental change in imputation quality. The inner procedure is performed until a \textit{selection stopping criterion} $\sigma$ is met, e.g. to make use of cached information from previous searches for heuristic speedup. Note that if we removed the model selection procedure from HyperImpute entirely, instead passing a fixed set of conditional specifications $\{a_{d}\}_{d=1}^{D}$ into the algorithm, we would recover conventional iterative imputation. As part of our experiments, we investigate and illustrate the sources of gain that stem from column-wise specification, automatic model selection, adaptive selection across iterations, as well as having a flexible catalogue of base learners (Section \ref{experiments}).
}

\begin{figure*}[ht!]
\vspace{0.25em}
\centering
\centerline{\includegraphics[width=\textwidth]{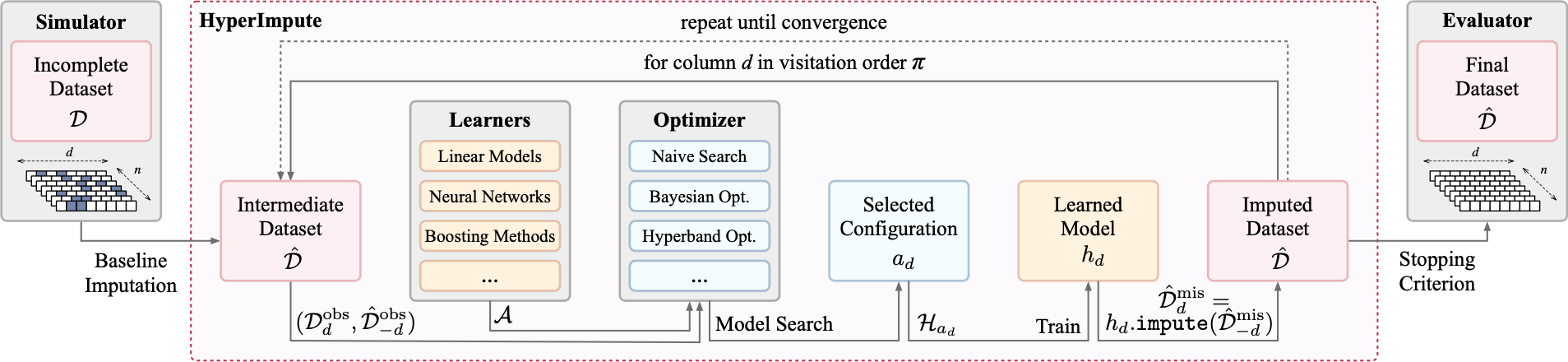}}
\vspace{-0.25em}
\caption{\squeeze{\textit{High-level overview of HyperImpute.} Blue indicates model selection algorithms and their selected output. Orange indicates candidate models and their trained output. Red indicates datasets and imputations. Gray indicates component modules in HyperImpute.}}
\label{hyperimpute:general_diagram}
\vspace{-0.75em}
\end{figure*}

\section{Practical Implementation}\label{method}

\squeeze{
In addition to the HyperImpute algorithm itself, our goal is also to facilitate accessibility and reproducibility in imputation research. Concretely, our implementation consists of:}
\vspace{-1em}
\begin{itemize}[leftmargin=1em]
\itemsep-0pt
\item \textbf{Learners}:
These are candidate classes for each univariate model, and include conditional specifications for both classification and regression---such as linear models, deep neural networks, and bagging and boosting methods. In Algorithm\pix\ref{alg:hyperimpute},\pix the global set of models and hyperparameters $\mathcal{A}$ includes the configuration space of all candidates selected by the user to be searched over for each variable. The ``plugin'' interface makes this trivially extensible: Additional learners simply need to conform to the \textit{fit-predict} paradigm and expose a well-defined hyperparameter set.
\item \squeeze{\textbf{Optimizers}:
These are candidate algorithms that implement the \texttt{ModelSearch} in Algorithm \ref{alg:hyperimpute}. Given an objective function, these focus on configuration \textit{selection} (e.g. bayesian optimization \cite{wang2017batched,alaa2018autoprognosis}), or on configuration \textit{evaluation} (e.g. adaptive computation \cite{krueger2015fast,li2018hyperband}). Among the implemented options, we default to our adaptation of \textit{Hyperband} \cite{li2018hyperband} that accommodates configuration spaces $\mathcal{A}$ spanning different learner classes, and to using totals of RMSE (continuous) or negative AUROC (categorical) as objective. As above, the interface is extensible as desired.}
\item \squeeze{\textbf{Imputers}:
These are candidate imputation methods that serve two purposes: First, any existing imputer can fulfill the \texttt{BaselineImpute} function in Algorithm~\ref{alg:hyperimpute} to seed $\hat{\mathcal{D}}$ for the first iteration; in our experiments, we default to mean substitution \cite{hawthorne2005imputing}. Second, any imputer constitutes a benchmark algorithm in performance comparison experiments---such as any method in Table \ref{tab:related} (see Section \ref{experiments}). Like above, any new imputer simply needs to conform to the \textit{fit-transform} paradigm. We have implemented comprehensive benchmark algorithms from recent literature.}
\end{itemize}
\vspace{-0.75em}
\squeeze{
Finally, we also implement modules and interfaces for \textbf{\textit{simulation}} of missing data (according to different missing mechanisms), \textbf{\textit{evaluation}} of imputed data (according to different performance metrics), and \textbf{\textit{comparison}} of imputation meth- ods via seeded and systematically cross-validated experiments. HyperImpute is implemented as an \texttt{sklearn} transformer, so it is fully compatible with \texttt{sklearn}-pipelines, and can be easily integrated as a component of an existing pipeline (e.g. for a downstream prediction task \cite{pedregosa2011scikit,alaa2018autoprognosis,lee2019temporal,jarrett2020clairvoyance}).}

\vspace{-0.75em}
\section{Empirical Investigation}\label{experiments}

\begin{figure*}[t!]
\vspace{0.25em}
\centering
\includegraphics[width=\linewidth]{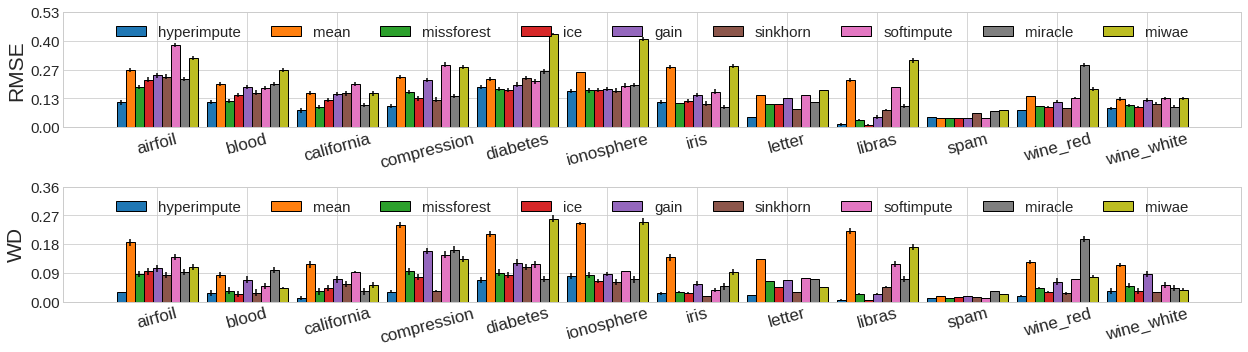}
\vskip -0.1in
\caption{\squeeze{\textit{Overall Performance.} Experiments on 12 UCI datasets under MAR at 0.3 missingness. Results shown as mean $\pm$ standard deviation of RMSE and WD. HyperImpute outperforms all benchmarks on both metrics in 10 of the 12, and at least one metric on all 12.}}
\label{fig:imputation_perf}
\vskip -0.05in
\end{figure*}

\begin{figure*}[t!]
\centering
\includegraphics[width=\linewidth]{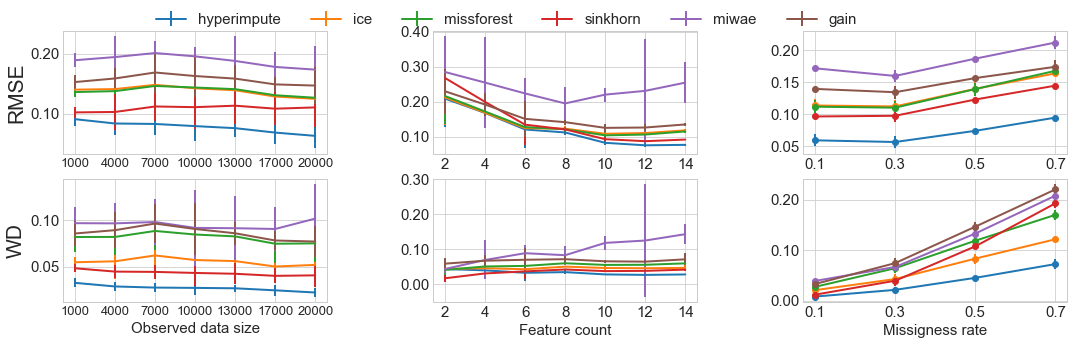}
\vskip -0.1in
\caption{\squeeze{\textit{Sensitivity Analysis.} Experiments performed on the \texttt{letter} dataset under the MAR setting. Results shown in terms of mean $\pm$ standard deviation of RMSE and WD, with sensitivities according to (a) observed data size, (b) feature count, and (c) missingness rate; when not perturbed for analysis, the observed data size is fixed at $N=$~20,000, the feature count at $D=$~14, and missingness rate at 0.3.}}
\label{fig:sensitivity_analysis}
\vskip -0.15in
\end{figure*}

\squeeze{Four aspects of HyperImpute deserve empirical investigation, and our goal in this section is to highlight them in turn:}
\vspace{-1em}
\begin{enumerate}[leftmargin=1.75em]
\itemsep-0pt
\item \squeeze{\textbf{Performance}: Bottom-line---\textit{Does HyperImpute work}? Section \ref{res1} compares the performance of HyperImpute with respect to a variety of state-of-the-art benchmarks.}
\item \squeeze{\textbf{Gains}: \textit{Why does it work?} Section \ref{res2} deconstructs various aspects of HyperImpute to investigate its sources of performance gain relative to classic iterative imputation.}
\item \squeeze{\textbf{Selection}: \textit{What does it learn?} Section \ref{res3} gives insight into the types of models that end up being selected, illustrating the process of adaptive and automatic selection.}
\item \squeeze{\textbf{Convergence}: \textit{Does HyperImpute converge?} Section \ref{res4} performs diagnostics on the iterative process of the method, illustrating its internal convergence behavior.}
\end{enumerate}
\vspace{-0.75em}
\squeeze{\textbf{Benchmarks}~
We test HyperImpute against the following:
%
Mean substitution, which imputes the column-wise unconditional mean (\textbf{Mean}) \cite{hawthorne2005imputing};
Imputation by chained equations, which is an iterative imputation method using linear/logistic models for conditional expectations (\textbf{ICE}); we follow the implementation in \cite{muzellec2020missing} using \cite{pedregosa2011scikit} based on \cite{van2011mice};
MissForest, a non-parametric iterative imputation algorithm using random forests as base learners (\textbf{MissForest}) \cite{stekhoven2012missforest};
Generative adversarial imputation networks, an adaptation of generative adversarial networks \cite{goodfellow2014generative,goodfellow2016nips} for missing data imputation, where the discriminator is now trained to classify the generator's output in an element-wise fashion (\textbf{GAIN}) \cite{yoon2018gain};
Missing data importance-weighted autoencoders, a deep latent variable model fit to missing data by optimizing a variational bound \cite{burda2015importance} adapted to the presence of missing data (\textbf{MIWAE}) \cite{mattei2019miwae}; 
SoftImpute, which performs imputation through soft-thresholded singular value decomposition, based on a low-rank assumption on the data (\textbf{SoftImpute}) \cite{hastie2015matrix};
Imputation models trained through optimal transport metrics, which leverages the assumption that two random batches of samples extracted from the same dataset should be similarly distributed, and uses Sinkhorn divergences between batches to quantify that objective (\textbf{Sinkhorn}) \cite{muzellec2020missing};
and a recent method using causal learning as a regularizer for progressive refinement of imputations by concurrently modeling the missing data mechanism itself (\textbf{MIRACLE}) \cite{kyono2021miracle}.
}

\textbf{Datasets}~
We employ 12 real-world datasets from the UCI machine learning repository \cite{dua2017uci}, similar to the experiment setup in recent works \cite{yoon2018gain,mattei2019miwae,muzellec2020missing}. To simulate \textbf{MCAR} data, the mask variable for each data point is realized according to a Bernoulli random variable with fixed mean. To simulate \textbf{MAR} data, a random subset of features is first set aside to be non-missing, and on the basis of which the remaining features are then masked: The masking mechanism takes the form of a logistic model that uses the non-missing features as inputs, and is parameterized by randomly chosen weights, with the bias term determined by the required rate of missingness. For completeness, we also simulate \textbf{MNAR} data for experiments---although this is not the focus of our work: This is either done by further masking the input features of the MAR mechanism according to a Bernoulli random variable with fixed mean, or by directly self-masking values using interval-censoring. In either MNAR mechanism, the missingness now depends on the missing values themselves.

\squeeze{
\textbf{Evaluation}~
Methods are evaluated according to how well the imputed values align with their ground-truth values, measured by the root-mean-square error (\textbf{RMSE}); as well as how well the imputed distribution matches that of the ground-truth distribution, measured by the Wasserstein distance (\textbf{WD}), similar to \cite{muzellec2020missing}. For each dataset, benchmark, and experiment setting, evaluations are performed using 10 different random seeds, and we report the mean and standard deviations of the resulting performance metrics. HyperImpute is trained to output the conditional expectation of missing values; we defer an investigation of multiple imputation to future work. Throughout our experiments, we perform various \textit{sensitivities} by assessing how relative performance varies according to the (\textbf{a}) number of 
samples used: ``observed data size''; (\textbf{b}) number of features present: ``feature count''; (\textbf{c}) proportion of missing values: ``missingness rate''; and (\textbf{d}) missingness mechanism: MCAR, MAR, MNAR.
The following subsections contain the most relevant results for the MAR setting; see Appendix \ref{appa} for further details on datasets and implementations, and see Appendix \ref{appb} for complete experiments, sensitivity analyses, and ablation studies.
}

\begin{table*}[tbp]
\renewcommand{\arraystretch}{0.955}
\centering
\vspace{0.25em}
\caption{\squeeze{\textit{Source of Gains.} Experiments under the MAR setting at a missingness rate of 0.3. Results shown in terms of mean $\pm$ standard deviation of RMSE for different sensitivities. See Table  \ref{tab:imputation_settings} for legend. All numbers are scaled by a factor of 10 for readability. Best is bold.}}
\vskip -0.075in
\label{tab:ablation_study_rmse}
\setlength{\tabcolsep}{10.8pt}
\renewcommand{\arraystretch}{0.90}
\begin{adjustbox}{max width=\textwidth}
\begin{tabular}{l|cccc|ccccc}
\toprule
\textit{Setting}              & \textbf{A} & \textbf{B} & \textbf{C} & \textbf{D} & \textit{airfoil} & \textit{california} & \textit{compression} & \textit{letter} & \textit{wine\_white}\\
\midrule
\texttt{ice\_lr}              & $\times$ & $\times$ & $\times$ & $\times$ & 2.349 $\pm$ 0.483 &  0.789 $\pm$ 0.324 &  1.429 $\pm$ 0.215 &  1.087 $\pm$ 0.149 &  0.919 $\pm$ 0.060 \\ 
\texttt{ice\_rf}              & $\times$ & $\times$ & $\times$ & $\times$ &  2.365 $\pm$ 0.533 &  0.777 $\pm$ 0.339 &  1.615 $\pm$ 0.141 &  1.118 $\pm$ 0.148 &  0.987 $\pm$ 0.034 \\ 
\texttt{ice\_cb}              & $\times$ & $\times$ & $\times$ & $\times$ &  2.078 $\pm$ 0.482 &  0.726 $\pm$ 0.327 &  1.251 $\pm$ 0.204 &  0.750 $\pm$ 0.099 &  0.877 $\pm$ 0.054 \\ 
\texttt{global\_search}       & $\times$ & $\checkmark$ & $\times$ & $\checkmark$ &  1.599 $\pm$ 0.322 &  0.745 $\pm$ 0.326 &  1.031 $\pm$ 0.175 &  0.527 $\pm$ 0.067 &  0.853 $\pm$ 0.049 \\ 
\texttt{column\_naive}        & $\checkmark$ & $\times$ & $\times$ & $\checkmark$ &  1.861 $\pm$ 0.446 &  0.713 $\pm$ 0.333 &  1.094 $\pm$ 0.167 &  0.525 $\pm$ 0.065 &  0.845 $\pm$ 0.081 \\ 
\texttt{wo\_flexibility\_rf}  & $\checkmark$ & $\checkmark$ & $\checkmark$ & $\times$ &  2.689 $\pm$ 0.755 &  0.762 $\pm$ 0.331 &  1.807 $\pm$ 0.186 &  1.117 $\pm$ 0.152 &  0.982 $\pm$ 0.031 \\ 
\texttt{wo\_flexibility\_cb}  & $\checkmark$ & $\checkmark$ & $\checkmark$ & $\times$ &  1.594 $\pm$ 0.367 &  0.740 $\pm$ 0.336 &  1.082 $\pm$ 0.157 &  0.757 $\pm$ 0.100 &  0.876 $\pm$ 0.050 \\ 
\texttt{wo\_adaptivity}       & $\checkmark$ & $\checkmark$ & $\times$ & $\checkmark$ &  1.665 $\pm$ 0.360 &  0.721 $\pm$ 0.354 &  1.047 $\pm$ 0.176 &  0.526 $\pm$ 0.066 &  0.890 $\pm$ 0.074 \\
\midrule
\texttt{HyperImpute}          & $\checkmark$ & $\checkmark$ & $\checkmark$ & $\checkmark$ &  \textbf{1.479} $\pm$ \textbf{0.294} &  \textbf{0.704} $\pm$ \textbf{0.336} &  \textbf{1.013} $\pm$ \textbf{0.145}  &  \textbf{0.524} $\pm$ \textbf{0.067}  &  \textbf{0.801} $\pm$ \textbf{0.053} \\ \bottomrule
\end{tabular}
\end{adjustbox}
\end{table*}

\begin{figure*}[tbp!]
\vskip -0.15in
\centering
\includegraphics[width=\linewidth,height=12em]{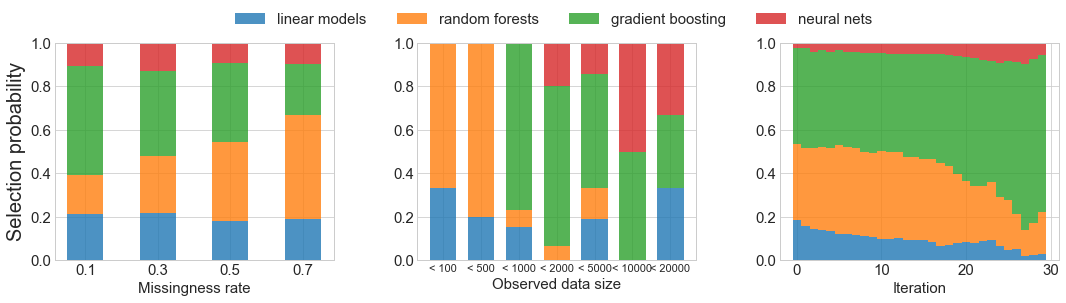}
\vskip -0.15in
\caption{\squeeze{\textit{Model Selections.} Experiments conducted under the MAR setting on 12 UCI datasets. Likelihood of different learner classes being selected for use as univariate models at various (a) missingness rates, (b) number of samples used, and (c) across iterations of the algorithm, with selection counts tallied across all columns and datasets; when not perturbed for analysis, the missingness rate is fixed at 0.3.}}
\label{fig:model_selection}
\vskip -0.15in
\end{figure*}

\vspace{-0.25em}
\subsection{Overall Performance}\label{res1}

\squeeze{
Figure \ref{fig:imputation_perf} shows the performance of HyperImpute and benchmark algorithms on all 12 datasets under the MAR setting at 30\% missingness rate. We observe that HyperImpute \textit{very consistently} performs at or above the level of all benchmarks: In particular, it outperforms all of them on 10 out of the 12 datasets with respect to both the RMSE and WD metrics. In Appendix \ref{appb}, we include much more comprehensive results collected in MCAR, MAR, and MNAR simulations at four levels of missingness rates, demonstrating that HyperImpute consistently performs better across a wide range of settings.}

\squeeze{
Moreover, to better evaluate HyperImpute's performance, we conduct a sensitivity analysis by varying the number of samples used, number of features present, and the missingness rate of the dataset. Figure \ref{fig:sensitivity_analysis} shows the performance of HyperImpute within these experiments against the five closest competitors (ICE, MissForest, GAIN, MIWAE, and Sinkhorn). Firstly, we see that as the number of samples increases, the performance improvement of HyperImpute relative to that of its benchmarks also increases. Secondly, we see that the advantage of HyperImpute is more noticeable with larger numbers of features (i.e. more than five); this is likely a byproduct of the iterative imputation scheme, since discriminative training of column-wise models become more challenging at lower-dimensions. Nonetheless, for feature counts above five, HyperImpute enjoys notable advantages. Thirdly, HyperImpute demonstrates significant improvement over benchmarks across the entire range of different rates of missingness: Importantly, it achieves low WD, suggesting it is less prone to overfitting to sparser datasets.}

\vspace{-0.25em}
\subsection{Source of Gains}\label{res2}

\squeeze{
HyperImpute is designed with a number of characteristics in mind (Section \ref{three}). Having empirically demonstrated strong overall results, an immediate question is how important these characteristics are for performance. Specifically, consider the source of gains from:
(\textbf{A}) \textit{column-wise} specification, (\textbf{B}) \textit{automatic} model and/or hyperparameter selection, (\textbf{C}) \textit{adaptive} selection across imputation iterations, and (\textbf{D}) having a \textit{flexible} catalogue of base learners. To disentangle the contributions of each of these to the final imputation performance of HyperImpute, here we deliberately ``switch off'' different properties and examine the resulting performance.
Table \ref{tab:imputation_settings} summarizes the possible combinations of settings. For settings \texttt{ice} and \texttt{wo\_flexibility}, we consider linear models, random forest, and catboost for the learner class.
}

\begin{table}[h!]
\setlength{\tabcolsep}{3pt}
\renewcommand{\arraystretch}{0.98}
\centering
\vspace{0.25em}
\caption{\textit{Legend for Source of Gains.} A ``learner class'' is a subset of $\mathcal{A}$ (e.g. all random forest models).
Note that ``{\scriptsize\texttt{wo\_adaptivity}}'' selects both models and hyperparameters, whereas ``{\scriptsize\texttt{column\_naive}}'' selects only the model class and uses its default hyperparameters.}
\vspace{-0.5em}
\begin{adjustbox}{max width=\linewidth}
\begin{tabular}{l|l|cccc}
\toprule
\textit{Description} & \textit{Name} & \textbf{A} & \textbf{B} & \textbf{C} & \textbf{D} \\ \midrule
\multirow{2}{*}{\textbf{ICE}} & \multirow{2}{*}{\texttt{ice}}              & \multirow{2}{*}{$\times$}      & \multirow{2}{*}{$\times$}      & \multirow{2}{*}{$\times$}      & \multirow{2}{*}{$\times$}      \\
&&&&& \\\midrule
\begin{tabular}[c]{@{}l@{}}Single model $a\in\mathcal{A}$ for all columns\\is auto-selected initially and then fixed\end{tabular}           & \texttt{global\_search}   & $\times$      & $\checkmark$  & $\times$      & $\checkmark$  \\ \midrule
\begin{tabular}[c]{@{}l@{}}Individual models $a_{d}\in\mathcal{A}$ for each column\\are auto-selected initially and then fixed\end{tabular}                                       & \texttt{column\_naive}    & $\checkmark$  & $\times$      & $\times$      & $\checkmark$  \\ \midrule
\begin{tabular}[c]{@{}l@{}}Like HyperImpute, but each column's\\model is limited to a single learner class\end{tabular}                            & \texttt{wo\_flexibility}  & $\checkmark$  & $\checkmark$  & $\checkmark$  & $\times$      \\ \midrule
\begin{tabular}[c]{@{}l@{}}Like HyperImpute, but each column's\\model is fixed after initial auto-selection\end{tabular}            & \texttt{wo\_adaptivity}   & $\checkmark$  & $\checkmark$  & $\times$      & $\checkmark$  \\ \midrule
\multirow{2}{*}{\textbf{HyperImpute}} & \multirow{2}{*}{\texttt{HyperImpute}}      & \multirow{2}{*}{$\checkmark$}  & \multirow{2}{*}{$\checkmark$}  & \multirow{2}{*}{$\checkmark$}  & \multirow{2}{*}{$\checkmark$}  \\
&&&&& \\
\bottomrule
\end{tabular}
\end{adjustbox}
\label{tab:imputation_settings}
\vspace{-1em}
\end{table}

\squeeze{
Results are shown in Table \ref{tab:ablation_study_rmse}. Observe that all four aspects of HyperImpute are crucial for good performance: Specifically, we note an average performance gain (i.e. decrease in RMSE) of 18\% as compared to the best \texttt{\small ice} models. Compared to results obtained when HyperImpute is run with a restricted class of base learners (\texttt{\small wo\_flexibility}), there is similarly an average of 11\% performance improvement. Compared to results obtained when models are only selected and applied globally (\texttt{\small global\_search}), there is an average of 4\% performance gain. Lastly, compared with naive column-wise selection (\texttt{\small column\_naive}) and the inclusion of a flexible pool of base learners without adaptive selection across iterations (\texttt{\small wo\_adaptivity}), HyperImpute sees an average of 7\% and 5\% performance gains respectively. WD results for different settings can be found in Appendix \ref{appb}.
}

\vspace{-0.35em}
\subsection{Model Selections}\label{res3}
\vspace{0.15em}

\squeeze{
Next, we illustrate what the HyperImpute algorithm tells us about different varieties of imputation problems. Across the same experimental settings and datasets as described in Section \ref{res1}, we investigate how the \textit{actual selections} for column-wise models differ across the missingness rates, the number of samples used, and across iterations of the algorithm. Figure \ref{fig:model_selection} summarizes the relative proportions of different classes of learners being selected across all experiments. Each time a class of learners is ultimately selected to impute a column within an iteration, it is included once within the tally. Interestingly, a diverse mix of models is selected across all experiments, and the composition of which varies systematically with the characteristics of the underlying datasets.}

\squeeze{
While varying the missingness rate, we observe that the empirical selection likelihood of neural networks (NN) and boosting (GB) decrease substantially with increasing missingness rates, whereas random forests (RF) and linear regressions (LR) become relatively more likely to be selected. This could reflect the tendency for highly expressive methods (NN, GB) to overfit when only given smaller rates of observable data, thus favoring simpler methods with lower-variance. We observe a similar trend when we vary dataset sizes---LR and RF appear to dominate when fewer than 500 samples are available; in contrast, NN and GB are substantially more commonly selected for larger datasets where more data is available to optimize their larger parameter sets.}

\squeeze{
Lastly, we study how adaptive model selection behaves across imputation iterations. As the number of iterations required for convergence varies by dataset, we let HyperImpute run for 30 iterations across all datasets to obtain comparable results. The changes in selection patterns indicate that HyperImpute indeed selects \textit{different types} of models across iterations as the baseline imputations get updated and improve over time. In particular, we observe that GB---and, to a lesser extent, NNs--are more commonly selected for later rounds when imputations stabilize, presumably because more emphasis can be placed on correctly modeling more difficult conditional imputations. In Appendix \ref{appendix:model_selection}, we present similar analyses for the MCAR and MNAR settings.}

\vspace{-0.25em}
\subsection{Convergence}\label{res4}
\vspace{0.15em}

\begin{figure}[t!]
\vspace{0.25em}
\makebox[\linewidth][c]{
\includegraphics[width=\linewidth]{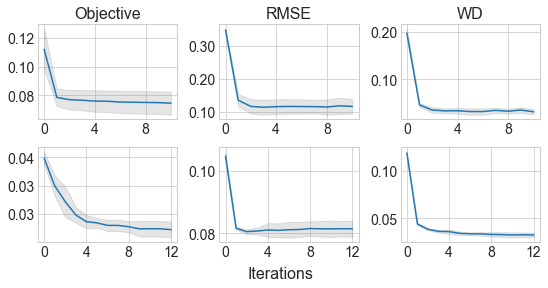}
}
\vskip -0.15in
\caption{\squeeze{\textit{Convergence.} Mean $\pm$ standard deviation of (a) the value of the objective function, (b) the RMSE metric, and (c) the WD metric, across iterations in experiments performed on \texttt{iris} (top) and \texttt{wine\_white} (bottom) datasets under the MAR setting and at 0.3 missingness. Note that iteration $0$ denotes results \textit{after} the initial round of mean imputations, and iteration $1$ denotes results \textit{after} an additional complete round-robin of conditional models are selected and trained by HyperImpute. We observe that HyperImpute significantly speeds up convergence within the iterative framework.}}
\label{fig:convergence_diagrams}
\vspace{-1em}
\end{figure}

Finally, we verify that HyperImpute successfully \textit{converges}: We compare the model's ``internal'' view of the imputation performance---measured by the value of the objective function during model selection, to the ground-truth imputation performance metrics (RMSE and WD). In Figure \ref{fig:convergence_diagrams}, we show results on representative datasets \texttt{iris} and \texttt{wine\_white}, with convergence results on additional datasets presented in Appendix \ref{appendix:convergence_results}. We observe that HyperImpute converges to a plateau quickly, generally within 4 iterations of the start, and that improvements in the model's internal objective correspond well to its ground-truth imputation performance.

\vspace{-0.35em}
\section{Conclusion}\label{conclusion}
\vspace{0.3em}

\squeeze{
Recent imputation research have often neglected iterative methods, relegating it to a trivial benchmarking exercise. To the contrary, our findings furnish a strong argument that a well-configured conditional specification easily produces state-of-the-art performance---an insight that may shape directions for future research. We introduced HyperImpute, a generalized iterative framework to automatically and adaptively configure column-wise models from expressive function approximators. We provide a practical implementation of the algorithm and comprehensive benchmarks, and an integrated suite of component tools for accessibility and reproducibility in imputation research. Finally, we demonstrated its use as an investigative platform for studying the characteristics and solutions to different imputation problems.}

\section*{Acknowledgments }

\squeeze{
We would like to thank all reviewers for all their invaluable feedback. This work was supported by AstraZeneca, Alzheimer's Research UK, the National Science Foundation (grant no. 1722516), and the US Office of Naval Research.}

\bibliography{hyperimpute,hyperimpute2}

\begin{thebibliography}{10}

\bibitem{barnard1999applications}
John Barnard and Xiao-Li Meng.
\newblock Applications of multiple imputation in medical studies: from aids to
  nhanes.
\newblock {\em Statistical methods in medical research}, 8(1):17--36, 1999.

\bibitem{mackinnon2010use}
A~Mackinnon.
\newblock The use and reporting of multiple imputation in medical research--a
  review.
\newblock {\em Journal of internal medicine}, 268(6):586--593, 2010.

\bibitem{stekhoven2012missforest}
Daniel~J Stekhoven and Peter B{\"u}hlmann.
\newblock Missforest—non-parametric missing value imputation for mixed-type
  data.
\newblock {\em Bioinformatics}, 28(1):112--118, 2012.

\bibitem{alaa2017personalized}
Ahmed~M Alaa, Jinsung Yoon, Scott Hu, and Mihaela Van~der Schaar.
\newblock Personalized risk scoring for critical care prognosis using mixtures
  of gaussian processes.
\newblock {\em IEEE Transactions on Biomedical Engineering}, 65(1):207--218,
  2017.

\bibitem{brand1999development}
Jaap Brand.
\newblock {\em Development, implementation and evaluation of multiple
  imputation strategies for the statistical analysis of incomplete data sets}.
\newblock 1999.

\bibitem{heckerman2000dependency}
David Heckerman, David~Maxwell Chickering, Christopher Meek, Robert
  Rounthwaite, and Carl Kadie.
\newblock Dependency networks for inference, collaborative filtering, and data
  visualization.
\newblock {\em Journal of Machine Learning Research}, 1(Oct):49--75, 2000.

\bibitem{raghunathan2001multivariate}
Trivellore~E Raghunathan, James~M Lepkowski, John Van~Hoewyk, Peter
  Solenberger, et~al.
\newblock A multivariate technique for multiply imputing missing values using a
  sequence of regression models.
\newblock {\em Survey methodology}, 27(1):85--96, 2001.

\bibitem{gelman2004parameterization}
Andrew Gelman.
\newblock Parameterization and bayesian modeling.
\newblock {\em Journal of the American Statistical Association},
  99(466):537--545, 2004.

\bibitem{van2006fully}
Stef Van~Buuren, Jaap~PL Brand, Catharina~GM Groothuis-Oudshoorn, and Donald~B
  Rubin.
\newblock Fully conditional specification in multivariate imputation.
\newblock {\em Journal of statistical computation and simulation},
  76(12):1049--1064, 2006.

\bibitem{van2011mice}
Stef Van~Buuren and Karin Groothuis-Oudshoorn.
\newblock mice: Multivariate imputation by chained equations in r.
\newblock {\em Journal of statistical software}, 45:1--67, 2011.

\bibitem{liu2014stationary}
Jingchen Liu, Andrew Gelman, Jennifer Hill, Yu-Sung Su, and Jonathan Kropko.
\newblock On the stationary distribution of iterative imputations.
\newblock {\em Biometrika}, 101(1):155--173, 2014.

\bibitem{zhu2015convergence}
Jian Zhu and Trivellore~E Raghunathan.
\newblock Convergence properties of a sequential regression multiple imputation
  algorithm.
\newblock {\em Journal of the American Statistical Association},
  110(511):1112--1124, 2015.

\bibitem{rezende2014stochastic}
Danilo~Jimenez Rezende, Shakir Mohamed, and Daan Wierstra.
\newblock Stochastic backpropagation and approximate inference in deep
  generative models.
\newblock In {\em International conference on machine learning}, pages
  1278--1286. PMLR, 2014.

\bibitem{gondara2017multiple}
Lovedeep Gondara and Ke~Wang.
\newblock Multiple imputation using deep denoising autoencoders.
\newblock {\em arXiv preprint arXiv:1705.02737}, 2017.

\bibitem{mattei2018leveraging}
Pierre-Alexandre Mattei and Jes Frellsen.
\newblock Leveraging the exact likelihood of deep latent variable models.
\newblock {\em arXiv preprint arXiv:1802.04826}, 2018.

\bibitem{yoon2018gain}
Jinsung Yoon, James Jordon, and Mihaela Schaar.
\newblock Gain: Missing data imputation using generative adversarial nets.
\newblock In {\em International Conference on Machine Learning}, pages
  5689--5698. PMLR, 2018.

\bibitem{li2019misgan}
Steven Cheng-Xian Li, Bo~Jiang, and Benjamin Marlin.
\newblock Misgan: Learning from incomplete data with generative adversarial
  networks.
\newblock {\em arXiv preprint arXiv:1902.09599}, 2019.

\bibitem{yoon2020gamin}
Seongwook Yoon and Sanghoon Sull.
\newblock Gamin: generative adversarial multiple imputation network for highly
  missing data.
\newblock In {\em Proceedings of the IEEE/CVF Conference on Computer Vision and
  Pattern Recognition}, pages 8456--8464, 2020.

\bibitem{nazabal2020handling}
Alfredo Nazabal, Pablo~M Olmos, Zoubin Ghahramani, and Isabel Valera.
\newblock Handling incomplete heterogeneous data using vaes.
\newblock {\em Pattern Recognition}, 107:107501, 2020.

\bibitem{ivanov2018variational}
Oleg Ivanov, Michael Figurnov, and Dmitry Vetrov.
\newblock Variational autoencoder with arbitrary conditioning.
\newblock {\em arXiv preprint arXiv:1806.02382}, 2018.

\bibitem{richardson2020mcflow}
Trevor~W Richardson, Wencheng Wu, Lei Lin, Beilei Xu, and Edgar~A Bernal.
\newblock Mcflow: Monte carlo flow models for data imputation.
\newblock In {\em Proceedings of the IEEE/CVF Conference on Computer Vision and
  Pattern Recognition}, pages 14205--14214, 2020.

\bibitem{mattei2019miwae}
Pierre-Alexandre Mattei and Jes Frellsen.
\newblock Miwae: Deep generative modelling and imputation of incomplete data
  sets.
\newblock In {\em International Conference on Machine Learning}, pages
  4413--4423. PMLR, 2019.

\bibitem{goodfellow2016deep}
Ian Goodfellow, Yoshua Bengio, and Aaron Courville.
\newblock {\em Deep learning}.
\newblock MIT press, 2016.

\bibitem{dai2021multiple}
Zongyu Dai, Zhiqi Bu, and Qi~Long.
\newblock Multiple imputation via generative adversarial network for
  high-dimensional blockwise missing value problems.
\newblock {\em arXiv preprint arXiv:2112.11507}, 2021.

\bibitem{salimans2016improved}
Tim Salimans, Ian Goodfellow, Wojciech Zaremba, Vicki Cheung, Alec Radford, and
  Xi~Chen.
\newblock Improved techniques for training gans.
\newblock {\em Advances in neural information processing systems},
  29:2234--2242, 2016.

\bibitem{thanh2020catastrophic}
Hoang Thanh-Tung and Truyen Tran.
\newblock Catastrophic forgetting and mode collapse in gans.
\newblock In {\em 2020 International Joint Conference on Neural Networks
  (IJCNN)}, pages 1--10. IEEE, 2020.

\bibitem{zhao2017towards}
Shengjia Zhao, Jiaming Song, and Stefano Ermon.
\newblock Towards deeper understanding of variational autoencoding models.
\newblock {\em arXiv preprint arXiv:1702.08658}, 2017.

\bibitem{lucas2019understanding}
James Lucas, George Tucker, Roger Grosse, and Mohammad Norouzi.
\newblock Understanding posterior collapse in generative latent variable
  models.
\newblock 2019.

\bibitem{muzellec2020missing}
Boris Muzellec, Julie Josse, Claire Boyer, and Marco Cuturi.
\newblock Missing data imputation using optimal transport.
\newblock In {\em International Conference on Machine Learning}, pages
  7130--7140. PMLR, 2020.

\bibitem{jager2021benchmark}
Sebastian J{\"a}ger, Arndt Allhorn, and Felix Bie{\ss}mann.
\newblock A benchmark for data imputation methods.
\newblock {\em Frontiers in big Data}, page~48, 2021.

\bibitem{hawthorne2005imputing}
Graeme Hawthorne, Graeme Hawthorne, and Peter Elliott.
\newblock Imputing cross-sectional missing data: comparison of common
  techniques.
\newblock {\em Australian \& New Zealand Journal of Psychiatry},
  39(7):583--590, 2005.

\bibitem{silva2011missing}
Esther-Lydia Silva-Ram{\'\i}rez, Rafael Pino-Mej{\'\i}as, Manuel
  L{\'o}pez-Coello, and Mar{\'\i}a-Dolores Cubiles-de-la Vega.
\newblock Missing value imputation on missing completely at random data using
  multilayer perceptrons.
\newblock {\em Neural Networks}, 24(1):121--129, 2011.

\bibitem{euredit2005interim}
Euredit.
\newblock Interim report on evaluation criteria for statistical editing and
  imputation.
\newblock {\em Euredit Project}, 3, 2005.

\bibitem{morvan2020neumiss}
Marine~Le Morvan, Julie Josse, Thomas Moreau, Erwan Scornet, and Ga{\"e}l
  Varoquaux.
\newblock Neumiss networks: differentiable programming for supervised learning
  with missing values.
\newblock {\em arXiv preprint arXiv:2007.01627}, 2020.

\bibitem{perez2022benchmarking}
Alexandre Perez-Lebel, Ga{\"e}l Varoquaux, Marine Le~Morvan, Julie Josse, and
  Jean-Baptiste Poline.
\newblock Benchmarking missing-values approaches for predictive models on
  health databases.
\newblock {\em GigaScience}, 2022.

\bibitem{ma2020vaem}
Chao Ma, Sebastian Tschiatschek, Jos{\'e}~Miguel Hern{\'a}ndez-Lobato, Richard
  Turner, and Cheng Zhang.
\newblock Vaem: a deep generative model for heterogeneous mixed type data.
\newblock {\em arXiv preprint arXiv:2006.11941}, 2020.

\bibitem{ma2018eddi}
Chao Ma, Sebastian Tschiatschek, Konstantina Palla, Jos{\'e}~Miguel
  Hern{\'a}ndez-Lobato, Sebastian Nowozin, and Cheng Zhang.
\newblock Eddi: Efficient dynamic discovery of high-value information with
  partial vae.
\newblock {\em arXiv preprint arXiv:1809.11142}, 2018.

\bibitem{lewis2021accurate}
Sarah Lewis, Tatiana Matejovicova, Yingzhen Li, Angus Lamb, Yordan Zaykov,
  Miltiadis Allamanis, and Cheng Zhang.
\newblock Accurate imputation and efficient data acquisitionwith
  transformer-based vaes.
\newblock In {\em NeurIPS 2021 Workshop on Deep Generative Models and
  Downstream Applications}, 2021.

\bibitem{morvan2021s}
Marine~Le Morvan, Julie Josse, Erwan Scornet, and Ga{\"e}l Varoquaux.
\newblock What's a good imputation to predict with missing values?
\newblock {\em arXiv preprint arXiv:2106.00311}, 2021.

\bibitem{rubin1976inference}
Donald~B Rubin.
\newblock Inference and missing data.
\newblock {\em Biometrika}, 63(3):581--592, 1976.

\bibitem{rubin1987multiple}
Donald~B Rubin.
\newblock {\em Multiple imputation for nonresponse in surveys}, volume~81.
\newblock John Wiley \& Sons, 1987.

\bibitem{van2018flexible}
Stef Van~Buuren.
\newblock {\em Flexible imputation of missing data}.
\newblock CRC press, 2018.

\bibitem{seaman2013meant}
Shaun Seaman, John Galati, Dan Jackson, and John Carlin.
\newblock What is meant by “missing at random”?
\newblock {\em Statistical Science}, 28(2):257--268, 2013.

\bibitem{jung2011latent}
Hyekyung Jung, Joseph~L Schafer, and Byungtae Seo.
\newblock A latent class selection model for nonignorably missing data.
\newblock {\em Computational statistics \& data analysis}, 55(1):802--812,
  2011.

\bibitem{little1993pattern}
Roderick~JA Little.
\newblock Pattern-mixture models for multivariate incomplete data.
\newblock {\em Journal of the American Statistical Association},
  88(421):125--134, 1993.

\bibitem{mohan2018estimation}
Karthika Mohan, Felix Thoemmes, and Judea Pearl.
\newblock Estimation with incomplete data: The linear case.
\newblock In {\em Proceedings of the International Joint Conferences on
  Artificial Intelligence Organization}, 2018.

\bibitem{sportisse2020estimation}
Aude Sportisse, Claire Boyer, and Julie Josses.
\newblock Estimation and imputation in probabilistic principal component
  analysis with missing not at random data.
\newblock {\em Advances in Neural Information Processing Systems}, 33, 2020.

\bibitem{kyono2021miracle}
Trent Kyono, Yao Zhang, Alexis Bellot, and Mihaela van~der Schaar.
\newblock Miracle: Causally-aware imputation via learning missing data
  mechanisms.
\newblock {\em Advances in Neural Information Processing Systems}, 34, 2021.

\bibitem{ipsen2020not}
Niels~Bruun Ipsen, Pierre-Alexandre Mattei, and Jes Frellsen.
\newblock not-miwae: Deep generative modelling with missing not at random data.
\newblock {\em arXiv preprint arXiv:2006.12871}, 2020.

\bibitem{ma2021identifiable}
Chao Ma and Cheng Zhang.
\newblock Identifiable generative models for missing not at random data
  imputation.
\newblock {\em Advances in Neural Information Processing Systems}, 34, 2021.

\bibitem{burda2015importance}
Yuri Burda, Roger Grosse, and Ruslan Salakhutdinov.
\newblock Importance weighted autoencoders.
\newblock {\em arXiv preprint arXiv:1509.00519}, 2015.

\bibitem{marker2002large}
David~A Marker, David~R Judkins, and Marianne Winglee.
\newblock Large-scale imputation for complex surveys.
\newblock {\em Survey nonresponse}, 329341, 2002.

\bibitem{troyanskaya2001missing}
Olga Troyanskaya, Michael Cantor, Gavin Sherlock, Pat Brown, Trevor Hastie,
  Robert Tibshirani, David Botstein, and Russ~B Altman.
\newblock Missing value estimation methods for dna microarrays.
\newblock {\em Bioinformatics}, 17(6):520--525, 2001.

\bibitem{garcia2010pattern}
Pedro~J Garc{\'\i}a-Laencina, Jos{\'e}-Luis Sancho-G{\'o}mez, and An{\'\i}bal~R
  Figueiras-Vidal.
\newblock Pattern classification with missing data: a review.
\newblock {\em Neural Computing and Applications}, 19(2):263--282, 2010.

\bibitem{hastie2015matrix}
Trevor Hastie, Rahul Mazumder, Jason~D Lee, and Reza Zadeh.
\newblock Matrix completion and low-rank svd via fast alternating least
  squares.
\newblock {\em The Journal of Machine Learning Research}, 16(1):3367--3402,
  2015.

\bibitem{little2002statistical}
Roderick~JA Little and Donald~B Rubin.
\newblock {\em Statistical analysis with missing data}, volume 793.
\newblock John Wiley \& Sons, 2002.

\bibitem{fitzmaurice2012applied}
Garrett~M Fitzmaurice, Nan~M Laird, and James~H Ware.
\newblock {\em Applied longitudinal analysis}, volume 998.
\newblock John Wiley \& Sons, 2012.

\bibitem{berti2014compatibility}
Patrizia Berti, Emanuela Dreassi, and Pietro Rigo.
\newblock Compatibility results for conditional distributions.
\newblock {\em Journal of Multivariate Analysis}, 125:190--203, 2014.

\bibitem{drechsler2008does}
J{\"o}rg Drechsler and Susanne R{\"a}ssler.
\newblock Does convergence really matter?
\newblock In {\em Recent advances in linear models and related areas}, pages
  341--355. Springer, 2008.

\bibitem{arnold2001conditionally}
Barry~C Arnold, Enrique Castillo, and Jose~Maria Sarabia.
\newblock Conditionally specified distributions: an introduction (with comments
  and a rejoinder by the authors).
\newblock {\em Statistical Science}, 16(3):249--274, 2001.

\bibitem{hutter2019automated}
Frank Hutter, Lars Kotthoff, and Joaquin Vanschoren.
\newblock {\em Automated machine learning: methods, systems, challenges}.
\newblock Springer Nature, 2019.

\bibitem{he2021automl}
Xin He, Kaiyong Zhao, and Xiaowen Chu.
\newblock Automl: A survey of the state-of-the-art.
\newblock {\em Knowledge-Based Systems}, 212:106622, 2021.

\bibitem{wang2017batched}
Zi~Wang, Chengtao Li, Stefanie Jegelka, and Pushmeet Kohli.
\newblock Batched high-dimensional bayesian optimization via structural kernel
  learning.
\newblock In {\em International Conference on Machine Learning}, pages
  3656--3664. PMLR, 2017.

\bibitem{alaa2018autoprognosis}
Ahmed Alaa and Mihaela Schaar.
\newblock Autoprognosis: Automated clinical prognostic modeling via bayesian
  optimization with structured kernel learning.
\newblock In {\em International conference on machine learning}, pages
  139--148. PMLR, 2018.

\bibitem{krueger2015fast}
Tammo Krueger, Danny Panknin, and Mikio~L Braun.
\newblock Fast cross-validation via sequential testing.
\newblock {\em J. Mach. Learn. Res.}, 16(1):1103--1155, 2015.

\bibitem{li2018hyperband}
Lisha Li, Kevin Jamieson, Giulia DeSalvo, Afshin Rostamizadeh, and Ameet
  Talwalkar.
\newblock Hyperband: A novel bandit-based approach to hyperparameter
  optimization, 2018.

\bibitem{pedregosa2011scikit}
Fabian Pedregosa, Ga{\"e}l Varoquaux, Alexandre Gramfort, Vincent Michel,
  Bertrand Thirion, Olivier Grisel, Mathieu Blondel, Peter Prettenhofer, Ron
  Weiss, Vincent Dubourg, et~al.
\newblock Scikit-learn: Machine learning in python.
\newblock {\em the Journal of machine Learning research}, 12:2825--2830, 2011.

\bibitem{lee2019temporal}
Changhee Lee, William Zame, Ahmed Alaa, and Mihaela Schaar.
\newblock Temporal quilting for survival analysis.
\newblock In {\em The 22nd international conference on artificial intelligence
  and statistics}, pages 596--605. PMLR, 2019.

\bibitem{jarrett2020clairvoyance}
Daniel Jarrett, Jinsung Yoon, Ioana Bica, Zhaozhi Qian, Ari Ercole, and Mihaela
  van~der Schaar.
\newblock Clairvoyance: A pipeline toolkit for medical time series.
\newblock In {\em International Conference on Learning Representations}, 2020.

\bibitem{goodfellow2014generative}
Ian Goodfellow, Jean Pouget-Abadie, Mehdi Mirza, Bing Xu, David Warde-Farley,
  Sherjil Ozair, Aaron Courville, and Yoshua Bengio.
\newblock Generative adversarial nets.
\newblock {\em Advances in neural information processing systems}, 27, 2014.

\bibitem{goodfellow2016nips}
Ian Goodfellow.
\newblock Generative adversarial networks.
\newblock {\em NIPS 2016 Tutorial}, 2016.

\bibitem{dua2017uci}
Dheeru Dua and Casey Graff.
\newblock Uci machine learning repository.
\newblock {\em University of California, Irvine, School of Information and
  Computer Sciences}, 2017.

\end{thebibliography}
\bibliographystyle{unsrt}

\newpage
\appendix
\onecolumn

\section{Experiment Details}\label{appa}

\squeeze{
In this section, we discuss further experimental details. We first give an overview of dataset details (Section \ref{app:a1}) and simulation details (Section \ref{app:a2}). Then we discuss the configuration space (Section \ref{app:a3}) and search strategies (Section \ref{app:a4}) for \texttt{ModelSearch}. Finally, we describe the termination criterion (Section \ref{app:a5}) and a clarification on data types (Section \ref{app:a6}).}

\subsection{Dataset Details}\label{app:a1}

\begin{table}[ht!]\small
\centering
\renewcommand{\arraystretch}{1.1}
\begin{tabular}{cccc}
\hline
\textbf{Dataset}                          & \textbf{Number of Instances} & \textbf{Number of Features} & \textbf{Experiment Name} \\ \hline
Airfoil Self-Noise Dataset               & 1503                         & 6                           & airfoil                   \\ \hline
Blood Transfusion Service Center Dataset & 748                          & 5                           & blood                     \\ \hline
California Housing Dataset               & 20640                       & 9                      & california               \\ \hline
Concrete Compressive Strength Dataset               & 1030                       & 9                      & compression               \\ \hline
Diabetes Dataset                         & 442                          & 10                          & diabetes                  \\ \hline
Ionosphere Dataset                       & 351                          & 34                          & ionosphere                \\ \hline
Iris Dataset                             & 150                          & 4                           & iris                      \\ \hline
Letter Recognition Dataset               & 20000                        & 16                          & letter                    \\ \hline
Libras Movement Dataset                  & 360                          & 91                          & libras                    \\ \hline
Spambase Dataset                         & 4601                         & 57                          & spam                      \\ \hline
Wine Quality Dataset(Red)                & 1599                         & 12                          & wine\_red                 \\ \hline
Wine Quality Dataset(White)              & 4898                         & 12                          & wine\_white               \\ \hline
\end{tabular}
\caption{Datasets for Evaluation.}
\label{hyperimpute_evaluation_datasets}
\vspace{-0.5em}
\end{table}

\subsection{Simulation Details}\label{app:a2}

The procedures for simulating missingness in each dataset are adapted directly from the experiment setup and implementation of \cite{muzellec2020missing}, and are exactly replicable using the source code.

\begin{itemize}
\item \textbf{MCAR}: Each value is removed according to  the realization of a Bernoulli random variable with a fixed parameter.
\item \textbf{MAR}: First, a subset of variables is randomly selected to be fully-observed, so only the remaining variables can have values that are missing. Second, these remaining variables have values removed according to a logistic model with random weights, using the fully-observed variables as regressors. The desired rate of missingness is achieved by adjusting the bias term.
\item \textbf{MNAR}: This is done by either further removing the values of the input features in the MAR mechanism above, or by directly removing values using interval-censoring.
\end{itemize}

In all three cases MCAR, MAR, and MNAR, experiments are performed using 10\%, 30\%, 50\%, and 70\% missingness.

\subsection{Configuration Space}\label{app:a3}

In Table \cref{tab:hyperimpute_hyperparams_domain}, we present the full configuration space (models and associated hyperparameter ranges) we consider for the column-wise model selection within HyperImpute. We use linear/logistic regressions and random forests as implemented in \texttt{sklearn} , XGBoost from the \texttt{xgboost} python package, catboost from the \texttt{catboost} python package and neural nets implemented using \texttt{pytorch}.

\subsection{Search Strategies}\label{app:a4}

\paragraph{Objective Function} Any ModelSearch strategy requires a column-wise objective function which is optimized through cross-validation when choosing the best conditional imputation model for a given column. In our implementation, we differentiate between column types; depending on the label type, the evaluation objective would be to minimize RMSE (continuous labels) or to maximize AUROC (categorical labels).

\paragraph{Search Strategies} To efficiently explore configuration spaces $\mathcal{A}$ of models (linear models, gradient boosting, random forests, neural nets etc.) with disjoint sets of hyperparameters, we implemented a number of search strategies that take as input $\mathcal{A}$ and a pre-defined objective function and output the best model found under computational constraints. 

~\vspace{-0.75em}

\begin{table}[H]\small
\renewcommand{\arraystretch}{1.1}
\centering
\begin{tabular}{lll}
\hline
\textbf{Model Class} & \textbf{Regression Task} & \textbf{Classification Task} \\ \hline
Linear Models & \begin{tabular}[c]{@{}l@{}}
- max\_iter $\in [100, 1000, 10000]$. \\
- solver $\in$ [``auto", ``svd", ``cholesky", ``lsqr", \\
``sparse\_cg", ``sag", ``saga"]
\end{tabular} & \begin{tabular}[c]{@{}l@{}}
-solver $\in$ [``newton-cg", ``lbfgs", ``sag", ``saga"]\\ 
- C $\in [1e-3, 1e-2]$\\ 
-multi\_class $\in$ [``auto", ``ovr", ``multinomial"]\\ 
- class\_weight $\in$ [``balanced", None]\end{tabular} \\ \hline
XGBoost & \begin{tabular}[c]{@{}l@{}}
- reg\_lambda $\in [1e-3, 10.0]$\\ 
- reg\_alpha $\in [1e-3, 10.0]$\\ 
- colsample\_bytree $\in [0.1, 0.9]$\\ 
- colsample\_bynode $\in [0.1, 0.9]$\\ 
- colsample\_bylevel $\in [0.1, 0.9]$\\ 
- subsample $\in [0.1, 0.9]$\\ 
- lr $\in [1e-4, 1e-3, 1e-2]$\\ 
- max\_depth $\in [2, 9]$\\ 
- n\_estimators $\in [10, 100]$\end{tabular} & 
\begin{tabular}[c]{@{}l@{}}
- reg\_lambda $\in [1e-3, 10.0]$\\ 
- reg\_alpha $\in [1e-3, 10.0]$\\ 
- colsample\_bytree $\in [0.1, 0.9]$\\ 
- colsample\_bynode $\in [0.1, 0.9]$\\ 
- colsample\_bylevel $\in [0.1, 0.9]$\\ 
- subsample $\in [0.1, 0.9]$\\ 
- lr $\in [1e-4, 1e-3, 1e-2]$\\ 
- max\_depth $\in [2, 9]$\\ 
- min\_child\_weight $\in [0, 300]$\\ 
- n\_estimators $\in [10, 100]$\\ 
- max\_bin $\in [256, 512]$\\ 
- booster $\in$ [``gbtree", ``gblinear", ``dart"]
\end{tabular} \\ \hline
CatBoost & 
\begin{tabular}[c]{@{}l@{}}
- depth $\in [1,5]$\\ 
- n\_estimators $\in [10, 100]$\\ 
- grow\_policy $\in$ [None,``Depthwise", \\ ``SymmetricTree",``Lossguide"]
\end{tabular} & 
\begin{tabular}[c]{@{}l@{}}
- depth $\in [1,5]$\\ 
- n\_estimators $\in [10, 100]$\\ 
- grow\_policy $\in$ [None,``Depthwise", \\ ``SymmetricTree",``Lossguide"]
\end{tabular} \\ \hline
Random Forest & \begin{tabular}[c]{@{}l@{}}
- criterion $\in$ [``mse", ``mae"]\\ 
- max\_features $\in$ [``auto", ``sqrt", ``log2"]\\ 
- min\_samples\_split $\in [2, 5, 10]$\\ 
- min\_samples\_leaf $\in [2, 5, 10]$\\ 
- max\_depth $\in [1, 4]$
\end{tabular} & \begin{tabular}[c]{@{}l@{}}
- criterion $\in$ [``gini", ``entropy"]\\ 
- max\_features $\in$ [``auto", ``sqrt", ``log2"]\\ 
- min\_samples\_split $\in [2, 5, 10]$\\ 
- min\_samples\_leaf $\in [2, 5, 10]$\\ 
- max\_depth $\in [1, 4]$\end{tabular} \\ \hline
Neural Nets & \begin{tabular}[c]{@{}l@{}}
- n\_layers\_hidden $\in [1, 2]$\\ 
- n\_units\_hidden $\in [10, 100]$\\ 
- lr $\in [1e-4, 1e-3]$\\ 
- weight\_decay $\in [1e-4, 1e-3]$\\ 
- dropout $\in [0., 0.2]$\\ 
- clipping\_value $\in [0, 1]$\end{tabular} & 
\begin{tabular}[c]{@{}l@{}}
- n\_layers\_hidden $\in [1, 2]$\\ 
- n\_units\_hidden $\in [10, 100]$\\ 
- lr $\in [1e-4, 1e-3]$\\ 
- weight\_decay $\in [1e-4, 1e-3]$\\ 
- dropout $\in [0., 0.2]$\\ 
- clipping\_value $\in [0, 1]$
\end{tabular} \\ \hline
\end{tabular}
\caption{Hyperparameter domain for each model class, and for each task type.}
\label{tab:hyperimpute_hyperparams_domain}
\end{table}

\begin{enumerate}
\item \textbf{Naive Search Strategy}
\begin{itemize}
\item Each model in the search pool is evaluated using its default hyperparameters, on the objective function.
\item The method returns the best model across evaluations.
\item \textbf{Pros:} Fast. \textbf{Cons:} No exploration of hyperparameters.
\end{itemize} 

\item \textbf{Bayesian Optimization Strategy (Model-Specific)}
\begin{itemize}
\item For each model in the search pool, we run a dedicated Bayesian Optimization (BO) call to suggest which hyperparameters to test to improve the objective function.
\item As a final model, we use the model and hyperparameters associated with the best score across all BO runs.
\item \textbf{Pros:} Good exploration. \textbf{Cons:} Slow, the Bayesian Optimization needs to be separately executed for each model in the pool.
\end{itemize}
\item \textbf{Adapted HyperBand Strategy}
\begin{itemize}
\item For each model in the search pool we define a special parameter, $iterations$, translated to epochs in linear/neural nets, or estimators in forests/gradient boosting.
\item In a preprocessing step, we learn a scaling mapping between the iterations of the different models, by evaluating each model for $[1, 5, 10]$ iterations, and evaluating their learning rate.
\item We apply the standard HyperBand \cite{li2018hyperband} search algorithm on the model search pool and the objective function while scaling the $iterations$ values using the mapping learned in the preprocessing step.
\item \textbf{Pros:} Good exploration/performance balance. \textbf{Cons:} The model performance mappings might be imprecise.
\end{itemize}
\end{enumerate}

\subsection{Termination Criterion}\label{app:a5}

The termination criterion $\gamma$ in Algorithm \ref{alg:hyperimpute} is met if: (1) the total number of iterations (i.e. loops over all columns $d\in\pi$) exceeds a pre-specified limit, or (2) changes in imputed values fall below a norm-based threshold (here we use the max norm), or (3) the optimization objective (i.e. the ``imputation quality'') stops improving over multiple consecutive rounds. For the exact implementation, these conditions are specified directly within {\small\texttt{plugin\_hyperimpute.py}} in the source code.

\subsection{Data Types}\label{app:a6}

Unlike most popular recent works that treat all inputs as real-valued (see e.g.
\cite{yoon2018gain,li2019misgan,yoon2020gamin,mattei2019miwae,muzellec2020missing,kyono2021miracle,hastie2015matrix}
), HyperImpute appropriately handles \textit{both} categorical and continuous variables.
Specifically, HyperImpute automatically (1) defines separate tasks for each variable (categorical/continuous), (2) maintains corresponding classes of candidates (classifiers/regressions), and (3) searches in their respective hyperparameter domains using distinct loss/objective functions. Specifically, see Table \ref{tab:hyperimpute_hyperparams_domain}.

\section{Additional Results}\label{appb}

For step-by-step experiment code for generating the following results in Sections \ref{appendix:imputation_performance}---\ref{appendix:convergence_results}, please refer to the corresponding notebooks in the {\small\texttt{experiments/}} directory in the source code.

\subsection{Overall Performance}\label{appendix:imputation_performance}

In this section, we provide additional results to highlight HyperImpute's imputation performance across a range of different missingness scenarios. To be exact, we report imputation performance on $12$ UCI datasets as measured by RMSE and WD across three missingness scenarios, \{MCAR, MAR, and MNAR\} and four missingness rates, \{$0.1$, $0.3$, $0.5$, $0.7$\}. The experiments are performed on the same datasets and compared to the same benchmarks using the procedures described in the experimental setup. 

\begin{figure}[h!]
\centering

\subfloat[MCAR-0.1]{%
  \includegraphics[width=.9\columnwidth]{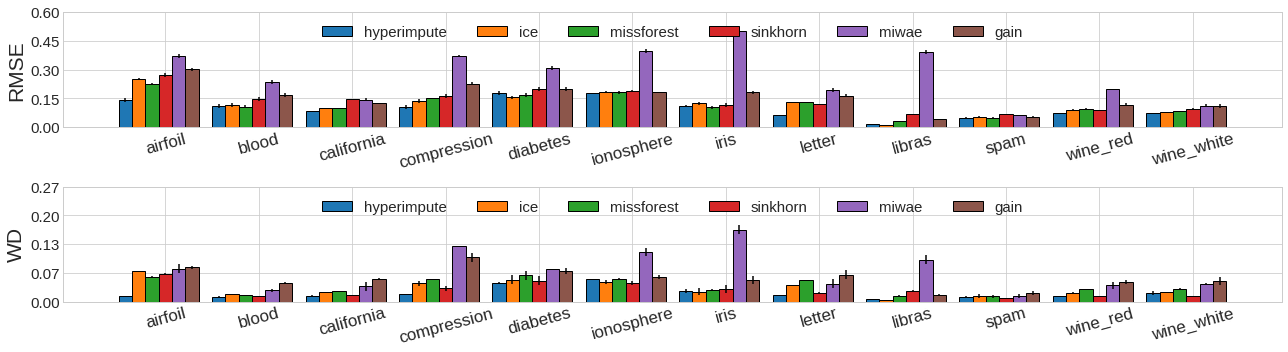}%
}

\subfloat[MCAR-0.3]{%
  \includegraphics[width=.9\columnwidth]{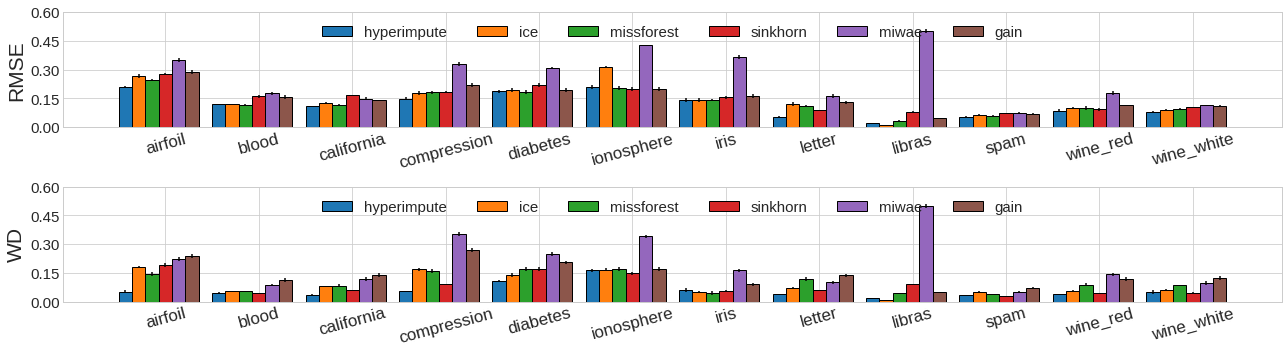}%
}

\subfloat[MCAR-0.5]{%
  \includegraphics[width=.9\columnwidth]{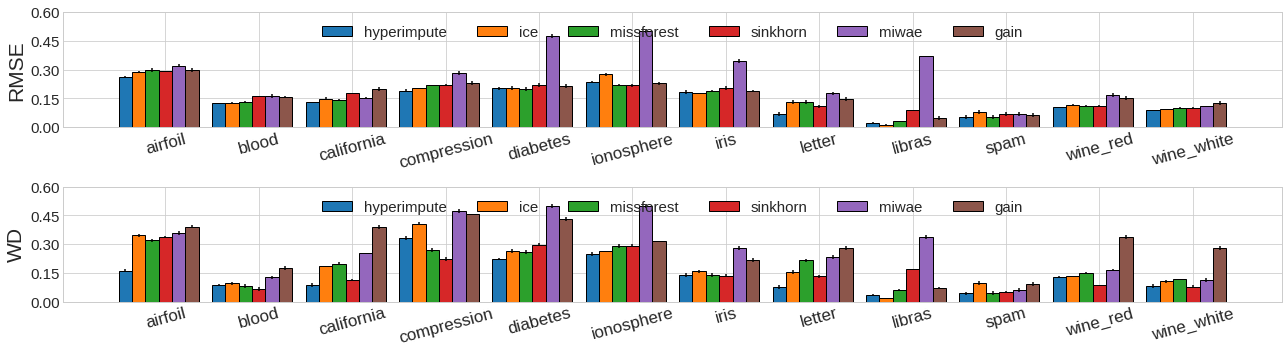}%
}

\subfloat[MCAR-0.7]{%
  \includegraphics[width=.9\columnwidth]{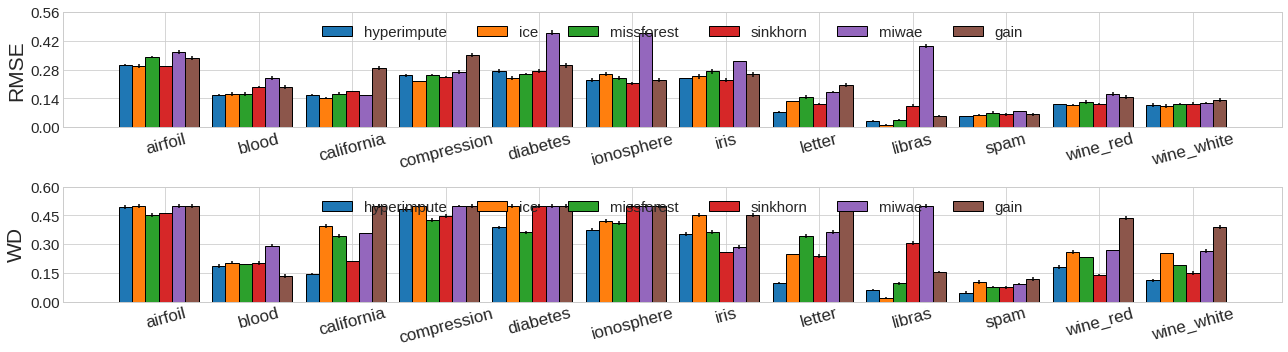}%
}

\caption{\textit{Overall Performance.} Experiments on $12$ UCI datasets under MCAR simulations at four levels of missingness---\{0.1, 0.3, 0.5, 0.7\}. Results shown as mean $\pm$ std of RMSE and WD.}
\label{fig:complete_mcar_results}
\end{figure}

\begin{figure}[h!]
\centering

\subfloat[MAR-0.1]{%
  \includegraphics[width=.9\columnwidth]{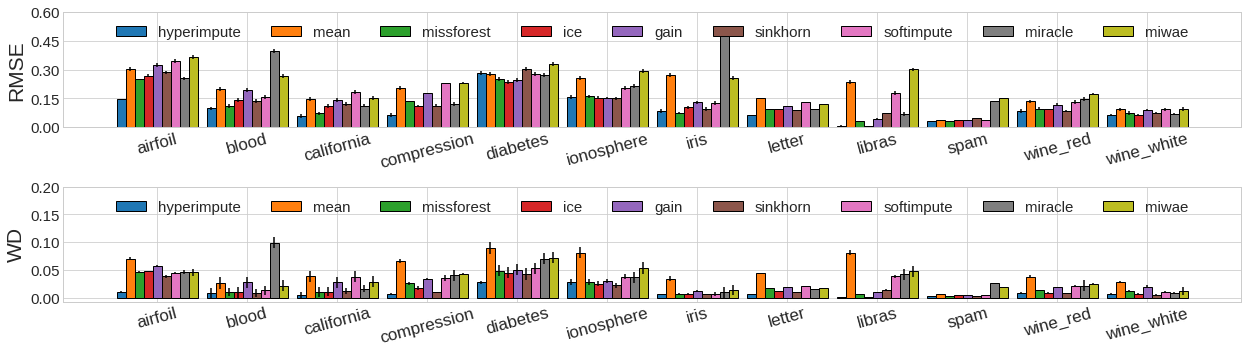}%
}

\subfloat[MAR-0.3]{%
  \includegraphics[width=.9\columnwidth]{img/model_predictiveness/general_overview_MAR_0.3.png}%
}

\subfloat[MAR-0.5]{%
  \includegraphics[width=.9\columnwidth]{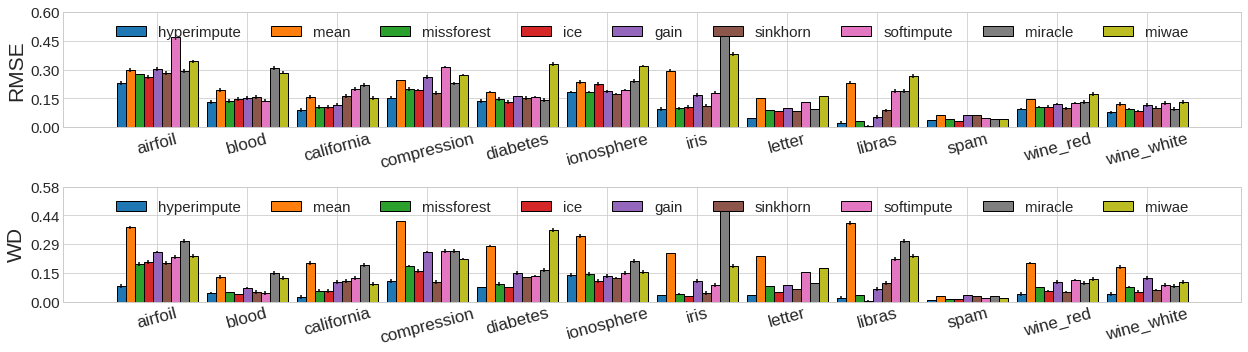}%
}

\subfloat[MAR-0.7]{%
  \includegraphics[width=.9\columnwidth]{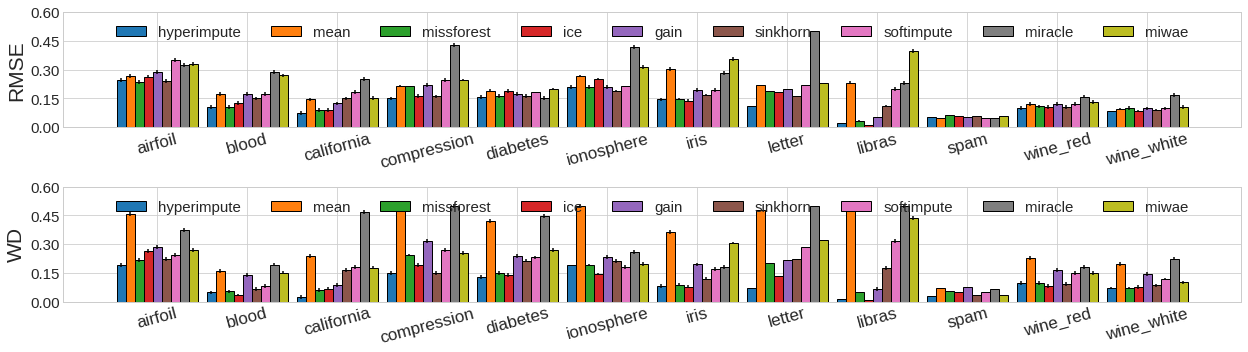}%
}

\caption{\textit{Overall Performance.} Experiments on $12$ UCI datasets under MAR simulations at four levels of missingness---\{0.1, 0.3, 0.5, 0.7\}. Results shown as mean $\pm$ std of RMSE and WD.}
\label{fig:complete_mar_results}
\end{figure}

\cref{fig:complete_mcar_results}, \cref{fig:complete_mar_results}, and \cref{fig:complete_mnar_results} plots the performance for MCAR, MAR, and MNAR respectively. Notably, HyperImpute out-performs the majority of benchmarks in terms of both RMSE and WD across different scenarios and missingness rates.

\subsection{Sensitivity Analysis}\label{appendix:sensitivity_analysis}

\begin{figure}[h!]
\centering

\subfloat[MCAR]{%
  \includegraphics[width=.9\columnwidth]{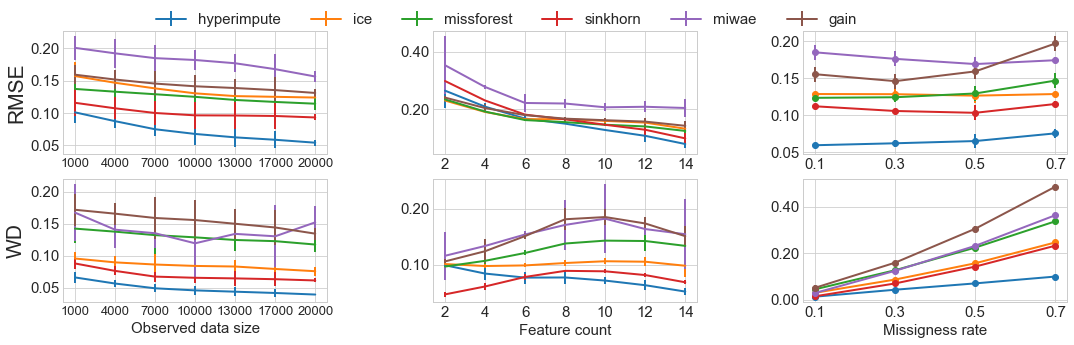}%
}

\subfloat[MAR]{%
  \includegraphics[width=.9\columnwidth]{img/model_perf/error_by_all_grouped_MAR.png}%
}

\subfloat[MNAR]{%
  \includegraphics[width=.9\columnwidth]{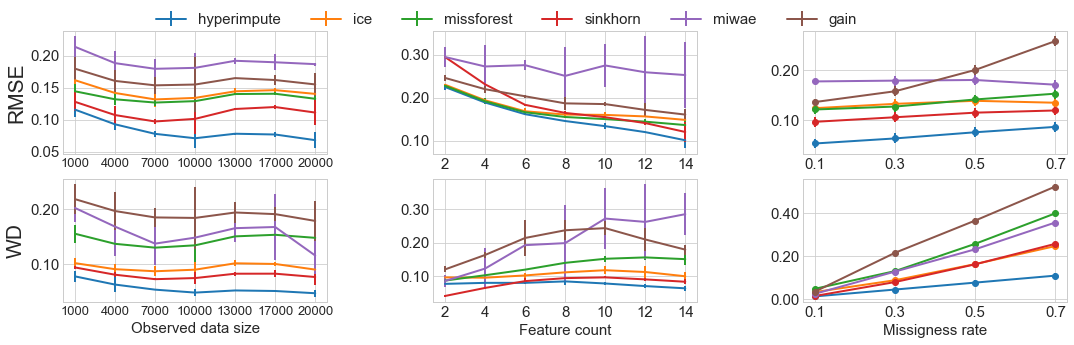}%
}

\caption{\textit{Sensitivity Analysis.} Experiments performed on the \texttt{letter} dataset under the MCAR, MAR and MNAR simulations. Results shown in terms of mean $\pm$ std of RMSE and WD with sensitivities according to (a) observed data size, (b) feature count, and (c) missingness rate. When not perturbed for analysis, the observed data size is fixed at $N=20,000$, feature count at $D=14$, and missingness rate at $0.3$.}
\label{fig:complete_sensitivity_analysis}
\end{figure}

Next, we quantitatively evaluate the robustness of HyperImpute to different missingness scenarios and missingness characteristics (i.e. observed data size, feature count, missingness rate) on the \texttt{letter} dataset. We perform sensitivity analysis by independently varying each of those parameters and plot the results in \cref{fig:complete_sensitivity_analysis}.

We see a few trends common across scenarios:
\begin{itemize}
    \item HyperImpute outperforms all benchmarks when lower number of samples are available, with the performance improvement more significant as data sizes increase,
    \item For settings with low feature counts, HyperImpute does not demonstrate markedly better imputation performance. This is likely due to the difficulty in the discriminatively training with less available regressors. However, at higher feature counts, HyperImpute demonstrates consistently superior performance,
    \item Lastly, HyperImpute achieves superior performance across all missingness scenarios and missingness rates. This advantage is more obvious at higher missingness rates, when performances of the benchmarks become significantly worse, but HyperImpute is able to minimise performance loss.
\end{itemize}

\subsection{Source of Gains}\label{appendix:ablation_study}

Here, we attach the complete results for our source of gains study, including RMSE and WD metrics on five UCI datasets. The RMSE scores for different settings are shown in \cref{tab:ablation_study_rmse_app} and WD scores in \cref{tab:ablation_study_wd_app}. We first note that all components of our algorithm, including (1) \textit{column-wise} imputation, (2) \textit{automatic} model and hyperparameter tuning, (3) \textit{adaptive} imputer selection across iterations, and (4) \textit{flexible} suite of base imputers, all contribute to performance improvements.

Specifically, we note an average performance gain of $18\%$ compared to the best \texttt{ice} models. Compared to results obtained when HyperImpute is run with a restricted class of base imputers, there is similarly a $11\%$ performance improvement. Additionally, in contrast to results obtained when an imputer is selected and applied globally, i.e. \texttt{global\_search}, there is a $4\%$ performance gain. Lastly, column-wise imputer selection and the inclusion of a flexible catalogue of base learners affords $7\%$ and $5\%$ performance gain, respectively.

Subsequently, we look at the distance between imputed data and the underlying ground truth. There is an average gain of $40\%$ in WD when compared to the best \texttt{ice} models. This becomes $27\%$ when we compare against the best results obtained using restricted base imputers, i.e. \texttt{wo\_flexibility}. In contrast to \texttt{global\_search}, the WDs are an average of $17\%$ lower. Lastly, column-wise imputer selection and flexible base learner classes present $14\%$ and $12\%$ additional performance improvement.

\begin{table}[h!]
\centering
\caption{\textit{Source of Gains.} Experiments under the MAR setting at a missingness rate of $0.3$. Results shown in terms of mean $\pm$ std of RMSE across different settings on $5$ datasets. All values are scaled by a factor of $10$ for readability. Best results are emboldened.}
\vskip -0.1in
\label{tab:ablation_study_rmse_app}
\setlength{\tabcolsep}{13.7pt}
\scalebox{.8}{
\begin{tabular}{c|ccccc}
\toprule
 & \textbf{airfoil} & \textbf{california} & \textbf{compression} & \textbf{letter} & \textbf{wine\_white}\\ \midrule
\texttt{ice\_lr}& $ 2.349 \pm 0.483 $& $ 0.789 \pm 0.324 $& $ 1.429 \pm 0.215 $& $ 1.087 \pm 0.149 $& $ 0.919 \pm 0.060 $\\ 
\texttt{ice\_rf}& $ 2.365 \pm 0.533 $& $ 0.777 \pm 0.339 $& $ 1.615 \pm 0.141 $& $ 1.118 \pm 0.148 $& $ 0.987 \pm 0.034 $\\ 
\texttt{ice\_cb}& $ 2.078 \pm 0.482 $& $ 0.726 \pm 0.327 $& $ 1.251 \pm 0.204 $& $ 0.750 \pm 0.099 $& $ 0.877 \pm 0.054 $\\ 
\texttt{global\_search}& $ 1.599 \pm 0.322 $& $ 0.745 \pm 0.326 $& $ 1.031 \pm 0.175 $& $ 0.527 \pm 0.067 $& $ 0.853 \pm 0.049 $\\ 
\texttt{column\_naive}& $ 1.861 \pm 0.446 $& $ 0.713 \pm 0.333 $& $ 1.094 \pm 0.167 $& $ 0.525 \pm 0.065 $& $ 0.845 \pm 0.081 $\\ 
\texttt{wo\_flexibility\_rf}& $ 2.689 \pm 0.755 $& $ 0.762 \pm 0.331 $& $ 1.807 \pm 0.186 $& $ 1.117 \pm 0.152 $& $ 0.982 \pm 0.031 $\\ 
\texttt{wo\_flexibility\_cb}& $ 1.594 \pm 0.367 $& $ 0.740 \pm 0.336 $& $ 1.082 \pm 0.157 $& $ 0.757 \pm 0.100 $& $ 0.876 \pm 0.050 $\\ 
\texttt{wo\_adaptivity}& $ 1.665 \pm 0.360 $& $ 0.721 \pm 0.354 $& $ 1.047 \pm 0.176 $& $ 0.526 \pm 0.066 $& $ 0.890 \pm 0.074 $\\ \hline
\texttt{HyperImpute}& $ \bm{1.479 \pm 0.294} $& $ \bm{0.704 \pm 0.336} $& $ \bm{1.013 \pm 0.145} $ & $ \bm{0.524 \pm 0.067} $ & $ \bm{0.801 \pm 0.053} $\\ \bottomrule
\end{tabular}}
\end{table}

\begin{table}[h!]
\centering
\caption{\textit{Source of Gains.} Experiments under the MAR setting at a missingness rate of $0.3$. Results shown in terms of mean $\pm$ std of WD across different settings on $5$ datasets. All values are scaled by a factor of $10$ for readability. Best results are emboldened.}
\vskip -0.1in
\label{tab:ablation_study_wd_app}
\setlength{\tabcolsep}{13.7pt}
\scalebox{.8}{
\begin{tabular}{c|ccccc}
\toprule
 & \textbf{airfoil} & \textbf{california} & \textbf{compression} & \textbf{letter} & \textbf{wine\_white}\\ \midrule
\texttt{ice\_lr}& $ 1.066 \pm 0.307 $& $ 0.330 \pm 0.192 $& $ 0.781 \pm 0.219 $& $ 0.526 \pm 0.069 $& $ 0.584 \pm 0.093 $\\ 
\texttt{ice\_rf}& $ 1.135 \pm 0.307 $& $ 0.354 \pm 0.173 $& $ 1.010 \pm 0.040 $& $ 0.799 \pm 0.106 $& $ 0.648 \pm 0.047 $\\ 
\texttt{ice\_cb}& $ 0.790 \pm 0.272 $& $ 0.196 \pm 0.107 $& $ 0.537 \pm 0.053 $& $ 0.400 \pm 0.045 $& $ 0.491 \pm 0.093 $\\ 
\texttt{global\_search}& $ 0.375 \pm 0.105 $& $ 0.253 \pm 0.171 $& $ 0.329 \pm 0.045 $& $ 0.257 \pm 0.010 $& $ 0.410 \pm 0.088 $\\ 
\texttt{column\_naive}& $ 0.516 \pm 0.278 $& $ 0.155 \pm 0.085 $& $ 0.315 \pm 0.030 $& $ 0.261 \pm 0.016 $& $ 0.390 \pm 0.172 $\\ 
\texttt{wo\_flexibility\_rf}& $ 1.217 \pm 0.366 $& $ 0.329 \pm 0.153 $& $ 1.103 \pm 0.079 $& $ 0.807 \pm 0.117 $& $ 0.645 \pm 0.060 $\\ 
\texttt{wo\_flexibility\_cb}& $ 0.443 \pm 0.142 $& $ 0.188 \pm 0.097 $& $ 0.369 \pm 0.057 $& $ 0.394 \pm 0.043 $& $ 0.495 \pm 0.096 $\\ 
\texttt{wo\_adaptivity}& $ 0.327 \pm 0.092 $& $ 0.170 \pm 0.111 $& $ 0.307 \pm 0.025 $& $ 0.256 \pm 0.009 $& $ 0.485 \pm 0.164 $\\ \hline
\texttt{HyperImpute}& $ \bm{0.291 \pm 0.103} $& $ \bm{0.114 \pm 0.060} $& $ \bm{0.289 \pm 0.041} $ & $ \bm{0.256 \pm 0.010} $ & $\bm{0.426 \pm 0.141}$\\ 
\bottomrule
\end{tabular}}
\end{table}

\subsection{Model Selections}\label{appendix:model_selection}

\begin{figure}[h!]
\centering

\subfloat[MCAR]{%
  \includegraphics[width=.9\columnwidth]{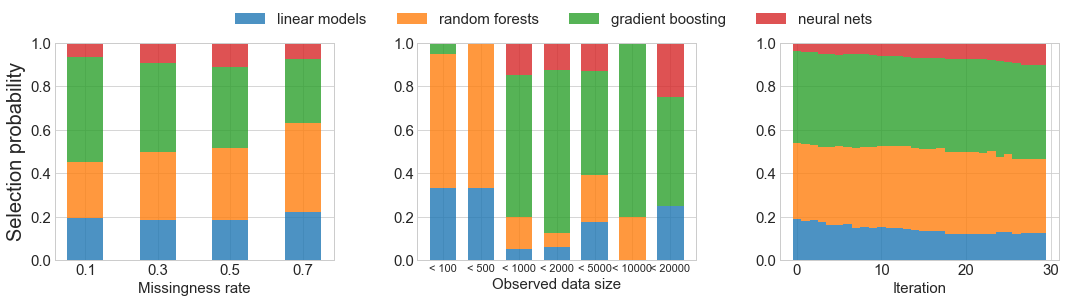}%
}

\subfloat[MAR]{%
  \includegraphics[width=.9\columnwidth]{img/model_sel/model_selection_MAR.png}%
}

\subfloat[MNAR]{%
  \includegraphics[width=.9\columnwidth]{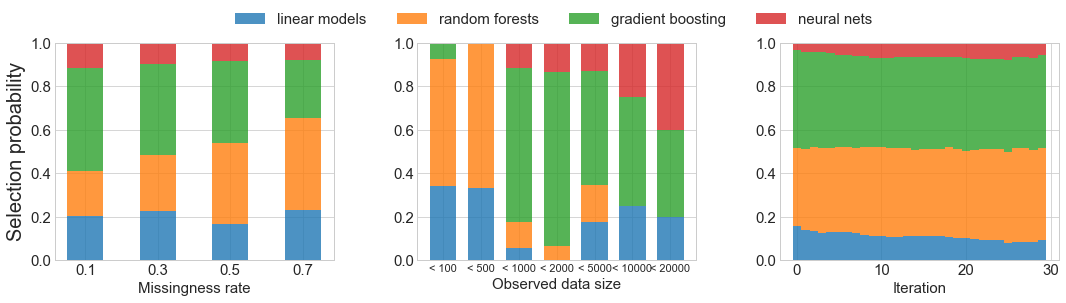}%
}

\caption{\textit{Model Selections.} Experiments conducted under the MCAR, MAR, and MNAR setting on $12$ UCI datasets. Likelihood of different learner classes being selected for use as univariate models at various (a) missingness rates, (b) number of samples used, and (c) across iterations of the algorithm, with selection counts tallied across all columns and datasets. When not perturbed for analysis, the missingness rate is fixed at $0.3$.}
\label{fig:complete_model_selection}
\end{figure}

We also wish to further investigate how the imputer selection changes under different missingness mechanisms. To do so, we tally the imputers chosen across columns, datasets and iterations. We plot how the likelihood of model selection varies with different missingness rates, number of samples and iterations in \cref{fig:complete_model_selection}. While missingness mechanisms differ, similar insights can be drawn.

Most notably, boosting algorithms and neural nets (NNs), which are highly expressive algorithms but require larger amounts of data to train, are more prevalent in low missingness rates and larger datasets. By contrast, linear regression (LR) and bagging algorithms, which are less expressive but lower variance predictors, are more common in low data regimes. We also note that HyperImpute tends to opt for more powerful algorithms in later iterations as it chooses to focus on more challenging imputations.

\subsection{Convergence}\label{appendix:convergence_results}

\begin{figure}[h!]
\centering

\subfloat[MNAR-0.1]{%
  \includegraphics[width=.9\columnwidth]{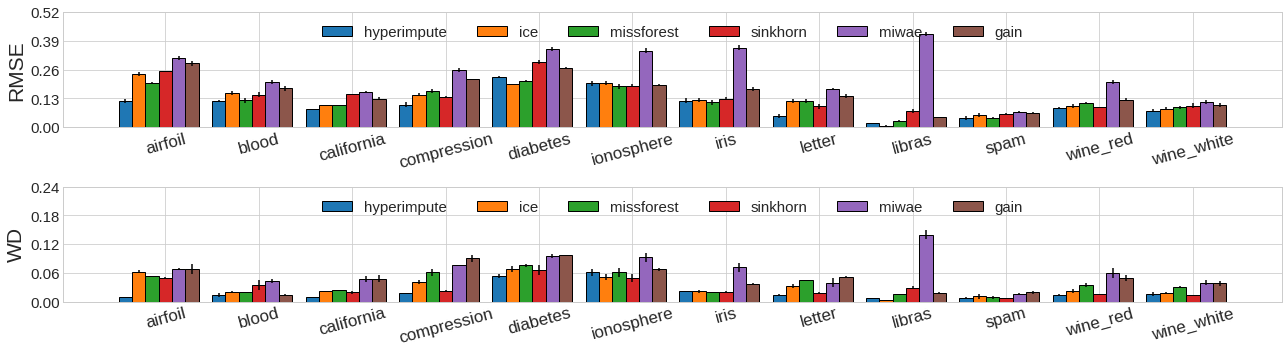}%
}

\subfloat[MNAR-0.3]{%
  \includegraphics[width=.9\columnwidth]{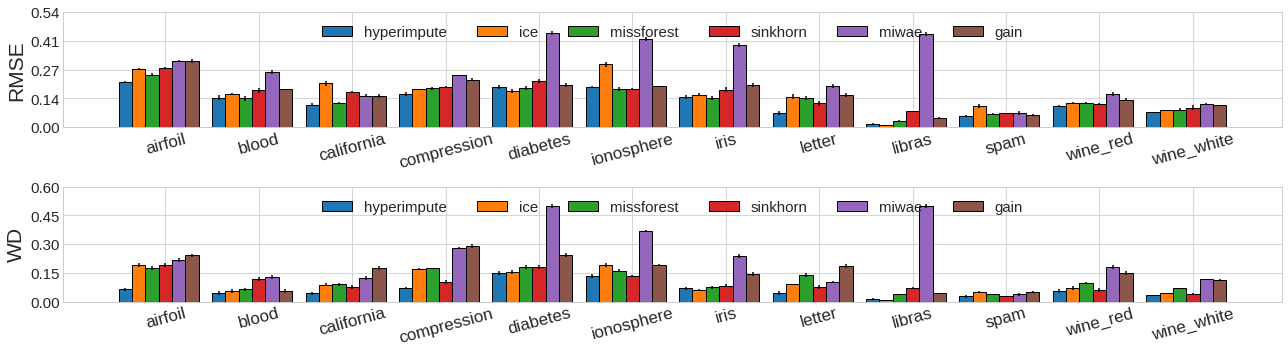}%
}

\subfloat[MNAR-0.5]{%
  \includegraphics[width=.9\columnwidth]{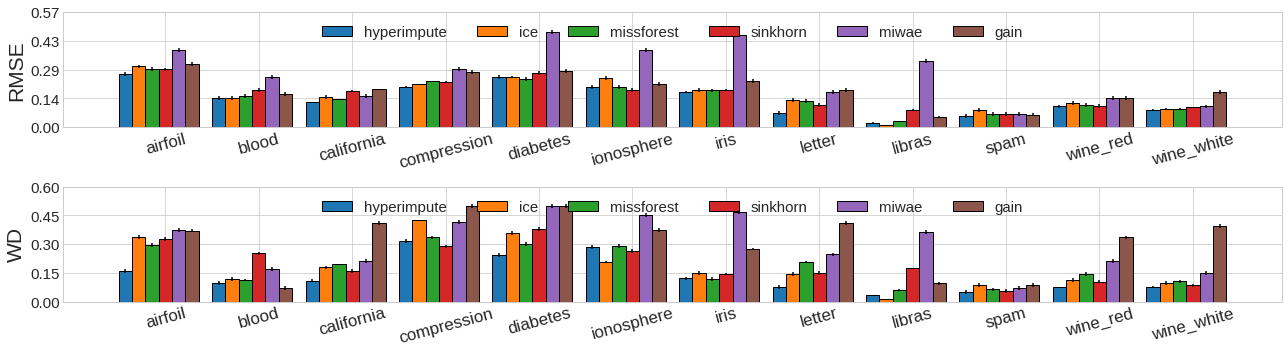}%
}

\subfloat[MNAR-0.7]{%
  \includegraphics[width=.9\columnwidth]{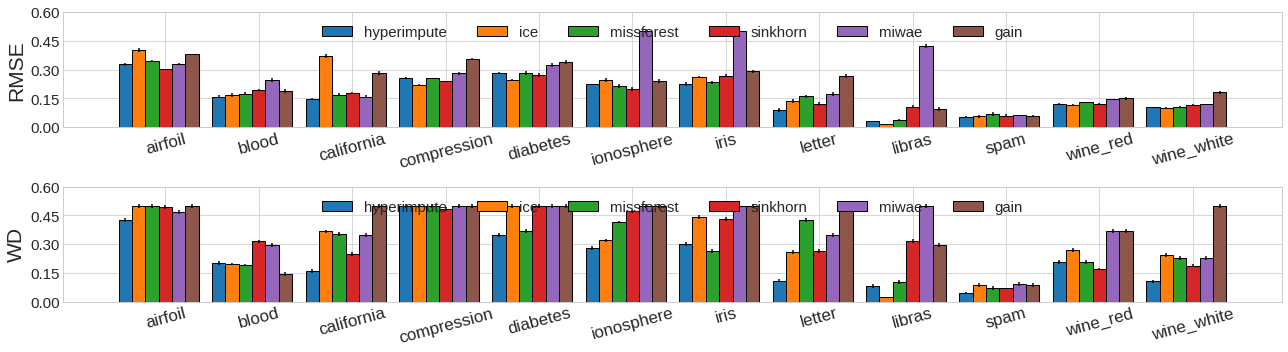}%
}

\caption{\textit{Overall Performance.} Experiments on $12$ UCI datasets under MNAR simulations at four levels of missingness---\{0.1, 0.3, 0.5, 0.7\}. Results shown as mean $\pm$ std of RMSE and WD.}
\label{fig:complete_mnar_results}
\end{figure}

Lastly, we present convergence results for MAR simulations at $0.3$ missingness for the $12$ UCI datasets employed in our experiments. We are interested in comparing the rates of convergence across imputation tasks. Evidently, convergence is generally achieved after $4$ iterations, with the model's internal objective corresponding well to ground-truth imputation performance. Of the $12$ experiments,  $3$ does not show converging behaviour even after $>10$ iterations. Firstly, we note that while the plots appear non-convergent, the actual fluctuations are very small due to the scale, i.e. $<0.01$ in RMSE. We additionally note that imputation performance is still superior to all benchmark methods in terms of both RMSE and WD on those datasets (see \cref{fig:complete_mnar_results}).

\begin{figure}[h!]
\centering
\includegraphics[width=.4\columnwidth]{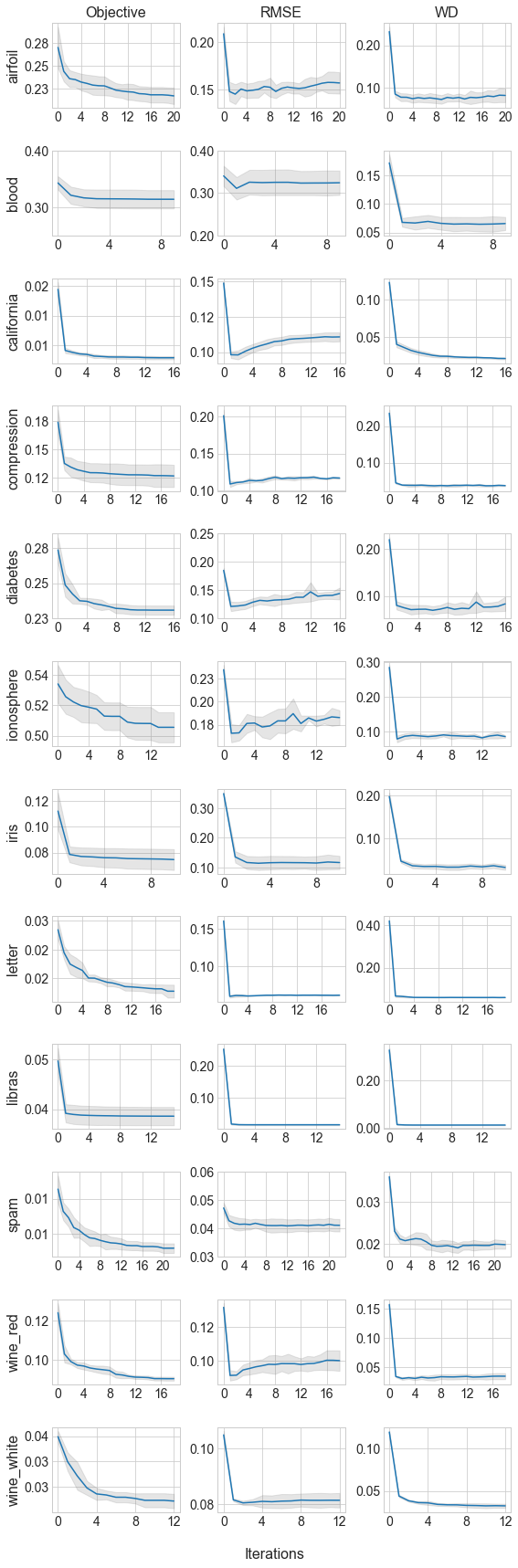}
\caption{\textit{Model Convergence.} Experiments performed using MAR at $0.3$ missingness rate on $12$ UCI datasets. Mean $\pm$ std of different metrics, including (a) objective error, (b) RMSE, (c) WD.}
\label{fig:complete_convergence_results}
\end{figure}

\clearpage

\section{Hyperparameters}

When speaking of ``hyperparameter optimization'', we must---importantly---first distinguish between (1) the hyperparameters of an \textbf{imputation method} (e.g. GAIN, MIWAE, Sinkhorn), or (2) the hyperparameters of each \textbf{column-wise model} (i.e. various regression/classification models) \textit{within} an iterative procedure (e.g. {ICE, MissForest, HyperImpute).

Regarding (1), since we work in the (realistic) setting where we are \textit{not} given access to completely-observed data during training,
it is theoretically \textbf{impossible} to perform hyperparameter optimization for an imputation method \textit{per se}:
To do so requires ``ground truths'' of the \textit{missing} values themselves (for measuring imputation quality), which we do not have.
(Note that cross-validating based on imputations of \textit{observed} values is futile: The identity $f(\mathcal{D})$$:=$$\mathcal{D}$
would appear globally optimal despite imputing nothing at all. And without knowing the true missingness pattern, naively adding \textit{artificial} missingness for the puposes of hyperparameter optimization would optimize for an incorrect objective).
In fact, GAIN, MIWAE, and Sinkhorn all operate in this setting: Their authors simply prescribe sensible defaults for hyperparameters---which we use.

Regarding (2), however, it is entirely \textbf{possible} to perform hyperparameter optimization for the column-wise models \textit{within} iterative imputation, using observed values (because the iterative procedure essentially reduces the original problem to a series of column-wise ``prediction'' problems): This is precisely what HyperImpute takes advantage of, and is what makes it a strict generalization of ICE, MissForest, or---for that matter---any iterative method that relies on a pre-selected set of conditional specifications. (Of course, in order to report final \textit{test-time} benchmarking results, we must employ non-missing held-out data for performance evaluation, but---again---the point is that such complete data is not available at \textit{training-time}).
}

Finally, note that HyperImpute is {\small\texttt{sklearn}}-compatible, and so it can be easily integrated as a component of an existing \texttt{sklearn}/AutoML pipeline (e.g. for a downstream prediction task \cite{pedregosa2011scikit,alaa2018autoprognosis,lee2019temporal,jarrett2020clairvoyance}).

\section{Running Time}

For some running time comparisons, see (left) figure below for an example on the {\small\texttt{spam}} dataset at various {\small MAR} missingness.
The main takeaway is that HyperImpute is far from being the most time-intensive. Interpreting wall-clock times requires care, but a key remark is that \textit{model training} tends to dominate (e.g. only using random forests often slows down MissForest), whereas HyperImpute's \textit{model selection} chooses/re-uses models from all classes---which can end up faster.
Laptop hardware: 32{\small GB RAM}, Intel Core i7-6700{\small HQ}, GeForce GTX 950{\small M}. All algorithms take order of seconds/minutes for convergence.

\begin{figure}[h!]
\centering
~~~~~~~~~~
\includegraphics[width=0.52\linewidth, trim=10em 0em 0em 0em]
{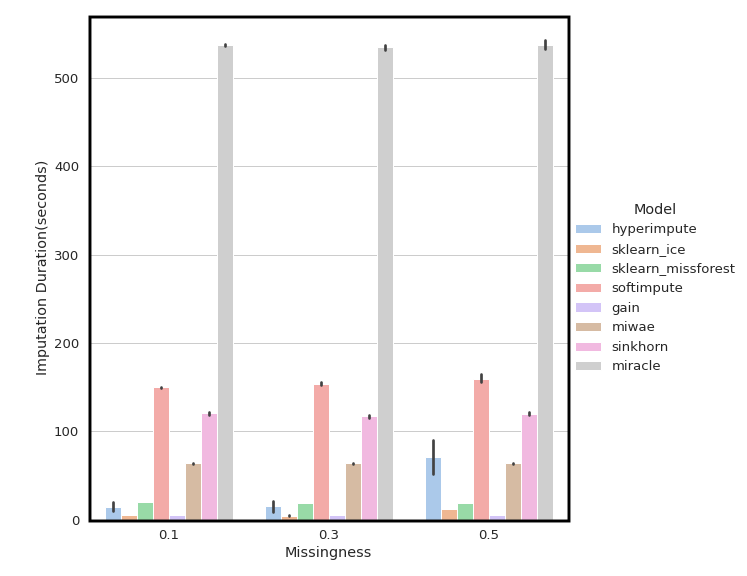}
~
\includegraphics[width=0.4225\linewidth, trim=7em 4.0em 3em 5.0em]
{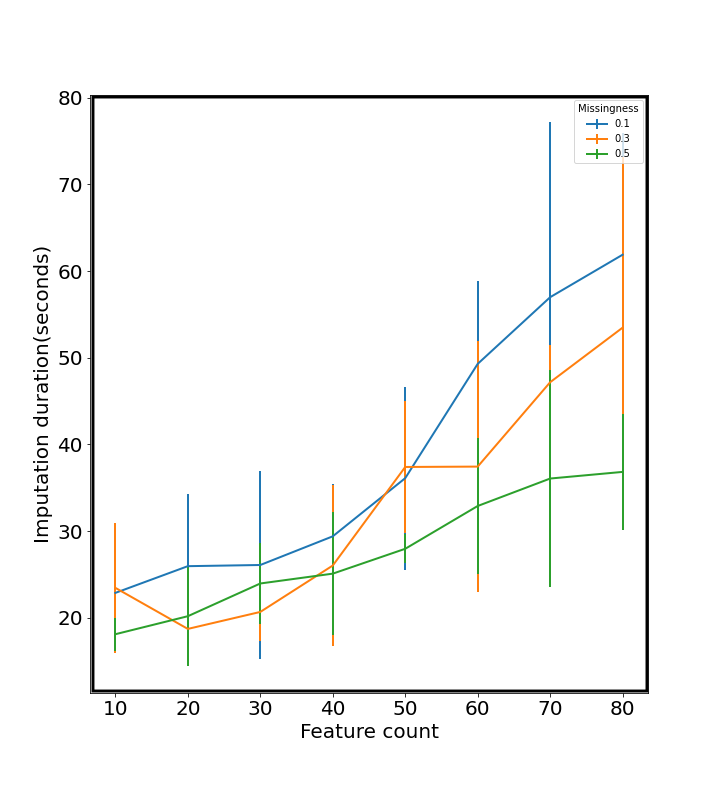}
\hspace{-1em}
\end{figure}

\squeeze{
In addition, we examine the effect of feature dimension on running time: See (right) figure above for an example on the largest dataset ({\footnotesize\texttt{libras}}) with various feature counts and missingness. In varying the feature count, features are subsetted from left to right in their original order of appearance in the raw dataset. Missingness is reported for 10\%, 30\%, and 50\%. Results are roughly consistent with the $O(D)$ complexity in number of features, and with our observation that model training dominates.
}

%
%
%
%
%
%
%

\end{document}